\documentclass{article}

\usepackage{microtype}
\usepackage{graphicx}
\usepackage{subfigure}
\usepackage{booktabs} \usepackage[dvipsnames]{xcolor}
\usepackage{colortbl}

\usepackage{hyperref}

\usepackage[accepted]{icml2025}

\usepackage{amsmath,amsfonts,bm}

\def\Secref#1{Sec.~\ref{#1}}

\def\eqref#1{eq.~\ref{#1}}
\def\Eqref#1{Eq.~\ref{#1}}

\def\1{\bm{1}}

\def\rd{{\textnormal{d}}}

\def\vx{{\bm{x}}}

\def\vz{{\bm{z}}}

\DeclareMathAlphabet{\mathsfit}{\encodingdefault}{\sfdefault}{m}{sl}
\SetMathAlphabet{\mathsfit}{bold}{\encodingdefault}{\sfdefault}{bx}{n}

\def\gD{{\mathcal{D}}}
\def\gE{{\mathcal{E}}}

\def\gG{{\mathcal{G}}}

\def\gL{{\mathcal{L}}}

\def\gU{{\mathcal{U}}}

\newcommand{\E}{\mathbb{E}}

\newcommand{\R}{\mathbb{R}}

 \usepackage{amssymb}
\usepackage{mathtools}
\usepackage{amsthm}
\usepackage{bm}
\usepackage{multirow,makecell}
\usepackage{pifont}
\usepackage{tikz}
\usetikzlibrary{arrows.meta,arrows,tikzmark}

\usepackage[capitalize,noabbrev]{cleveref}

\theoremstyle{plain}

\theoremstyle{definition}

\theoremstyle{remark}

\usepackage[textsize=tiny]{todonotes}

\usepackage{xspace}
\newcommand{\OURS}{$\epsilon$-VAE\xspace}

\newcommand{\ie}{\textit{i.e.}\xspace}
\newcommand{\eg}{\textit{e.g.}\xspace}

\newcommand{\etc}{\textit{etc.}\xspace}
\newcommand{\myparagraph}[1]{\textbf{#1}}

\newcommand{\cmark}{\ding{51}}
\newcommand{\xmark}{\ding{55}}
\newcommand{\bluecell}{\cellcolor{TealBlue!15}}
\newcommand{\redcell}{\cellcolor{WildStrawberry!15}}
\newcommand{\greencell}{\cellcolor{LimeGreen!15}}

\Crefname{figure}{Fig.}{Figs.}
\Crefname{table}{Tab.}{Tabs.}
\Crefname{section}{Sec.}{Secs.}
\Crefname{appendix}{Appx.}{Appxes.}

\icmltitlerunning{Epsilon-VAE: Denoising as Visual Decoding}

\begin{document}

\twocolumn[
\icmltitle{Epsilon-VAE: Denoising as Visual Decoding}

\icmlsetsymbol{equal}{*}

\begin{icmlauthorlist}
\icmlauthor{Long Zhao}{gdm}
\icmlauthor{Sanghyun Woo}{gdm}
\icmlauthor{Ziyu Wan}{gdm,city,equal}
\icmlauthor{Yandong Li}{gdm}
\icmlauthor{Han Zhang}{gdm}
\icmlauthor{Boqing Gong}{gdm}
\icmlauthor{Hartwig Adam}{gdm}
\icmlauthor{Xuhui Jia}{gdm}
\icmlauthor{Ting Liu}{gdm}
\end{icmlauthorlist}

\icmlaffiliation{gdm}{Google DeepMind}
\icmlaffiliation{city}{City University of Hong Kong}

\icmlcorrespondingauthor{Long Zhao}{longzh@google.com}
\icmlcorrespondingauthor{Xuhui Jia}{xhjia@google.com}
\icmlcorrespondingauthor{Ting Liu}{liuti@google.com}

\icmlkeywords{Diffusion Model, VAE, Image Tokenizer, Rectified Flow}

\vskip 0.3in
]

\printAffiliationsAndNotice{$^*$Work done as a student researcher at Google.} 

\begin{abstract}

In generative modeling, tokenization simplifies complex data into compact, structured representations, creating a more efficient, learnable space. For high-dimensional visual data, it reduces redundancy and emphasizes key features for high-quality generation. Current visual tokenization methods rely on a traditional autoencoder framework, where the encoder compresses data into latent representations, and the decoder reconstructs the original input. In this work, we offer a new perspective by proposing \textit{denoising as decoding}, shifting from single-step reconstruction to iterative refinement. Specifically, we replace the decoder with a diffusion process that iteratively refines noise to recover the original image, guided by the latents provided by the encoder. We evaluate our approach by assessing both reconstruction (rFID) and generation quality (FID), comparing it to state-of-the-art autoencoding approaches. By adopting iterative reconstruction through diffusion, our autoencoder, namely \OURS, achieves high reconstruction quality, which in turn enhances downstream generation quality by 22\% at the same compression rates or provides 2.3$\times$ inference speedup through increasing compression rates. We hope this work offers new insights into integrating iterative generation and autoencoding for improved compression and generation.

\end{abstract} \section{Introduction}

Two dominant paradigms in modern visual generative modeling are autoregression~\citep{radford2018improving} and diffusion~\citep{ho2020denoising}.
Tokenization is essential for both: discrete tokens allow step-by-step conditional generation in autoregressive models, while continuous latents enable efficient learning in the denoising process of diffusion models.
In either case, empirical results demonstrate that tokenization enhances generative performance.
Here, we focus on \textit{continuous} tokenization for latent diffusion models, which excel at generating high-dimensional visual data.

In this paper, we revisit the conventional autoencoding pipeline, which typically consists of an encoder that compresses the input into a latent representation and a decoder that reconstructs the original data in a single step. 
Instead of a deterministic decoder, we introduce a diffusion process~\citep{ho2020denoising,song2020score}, where the encoder still compresses the input into a latent representation, but reconstruction is performed iteratively through denoising. 
This reframing turns the reconstruction phase into a progressive refinement process, where the diffusion model, guided by the latent representation, gradually restores the original data.
While previous work~\cite{preechakul2022diffusion} and concurrent work~\cite{birodkar2024sample} have explored diffusion mechanisms in autoencoding, none have fully realized a practical diffusion-based autoencoder. By carefully co-designing architecture and objectives, we firstly show that our approach outperforms state-of-the-art autoencoding paradigms in reconstruction fidelity, sampling efficiency, and resolution generalization.

To effectively implement our approach, several key design factors must be carefully considered. 
First, the architectural design must ensure that the diffusion decoder is effectively conditioned on the encoder latent representations.
Second, the training objectives should leverage synergies with traditional autoencoding losses, such as LPIPS~\citep{zhang2018unreasonable} and GAN~\citep{esser2021taming}. Finally, diffusion-specific design choices play a crucial role, including: (1) model parameterization, which defines the prediction target for the diffusion decoder; (2) noise scheduling, which shapes the optimization trajectory; and (3) the distribution of time steps during training and testing, which balances noise levels for effective learning and generation.
Our study systematically examines these components through controlled experiments, demonstrating their impact on achieving a high-performing diffusion-based autoencoder. We show in the experiments that under the standard configuration~\citep{rombach2022high}, our method obtains a 40\% improvement in terms of reconstruction quality, leading to 22\% better image generation quality. More notably, we achieve 2.3$\times$ higher inference throughput by increasing compression rates, while keeping competitive generation quality.

In summary, our contributions are as follows: (1) introducing a novel approach that fully leverages the capabilities of diffusion decoders for more practical diffusion-based autoencoding, achieving strong rFID, high sampling efficiency (within 1 to 3 steps), and robust resolution generalization; (2) presenting key design choices in both architecture and objectives to optimize performance; and (3) conducting extensive controlled experiments that demonstrate our method achieves high-quality reconstruction and generation results, outperforming leading visual auto-encoding paradigms.

 \section{Background}

We start by briefly reviewing the basic concepts required to understand the proposed method. A more detailed summary of related work is deferred to \cref{sec:relatedwork}.

\myparagraph{Visual autoencoding.} 
To achieve efficient and scalable high-resolution image synthesis, common generative models, including autoregressive models~\citep{razavi2019generating,esser2021taming,chang2022maskgit} and diffusion models~\citep{rombach2022high}, are typically trained in a low-resolution latent space by first downsampling the input image using a tokenizer.
The tokenizer is generally implemented as a convolutional autoencoder consisting of an encoder, $\gE$, and a decoder, $\gG$. 
Specifically, the encoder, $\gE$, compresses an input image $\vx \in \R^{H \times W \times 3}$ into a set of latent codes (\ie, tokens), $\gE(\vx) =\vz \in \R^{H/f \times W/f \times n_z}$, where $f$ is the downsampling factor and $n_z$ is the latent channel dimensions. 
The decoder, $\gG$, then reconstructs the input from $\vz$, such that $\gG(\vz) = \vx$.

Training an autoencoder primarily involves several losses: reconstruction loss $\gL_\text{rec}$, perceptual loss (LPIPS) $\gL_\text{LPIPS}$, and adversarial loss $\gL_\text{adv}$.
The reconstruction loss minimizes pixel differences (\ie, typically measured by the $\ell_1$ or $\ell_2$ distance) between $\vx$ and $\gG(\vz)$. 
The LPIPS loss~\citep{zhang2018unreasonable} enforces high-level structural similarities between inputs and reconstructions by minimizing differences in their intermediate features extracted from a pre-trained VGG network~\citep{simonyan2015very}.
The adversarial loss~\citep{esser2021taming} introduces a discriminator, $\gD$, which encourages more photorealistic outputs by distinguishing between real images, $\gD(\vx)$, and reconstructions, $\gD(\gG(\vz))$. 
The final training objective is a weighted combination of these losses:
\begin{equation}
\label{eq:vqgan}
    \gL_\text{VAE} = \gL_\text{rec} + \lambda_\text{LPIPS}\cdot\gL_\text{LPIPS} + \lambda_\text{adv}\cdot\gL_\text{adv},
\end{equation}
where the $\lambda$ values are weighting coefficients. 
In this paper, we consider the autoencoder optimized by \Eqref{eq:vqgan} as our main competing baseline~\citep{esser2021taming}, as it has become a standard tokenizer training scheme widely adopted in state-of-the-art image and video generative models~\citep{chang2022maskgit,rombach2022high,yu2022vector,yu2023magvit,kondratyuk2024videopoet,esser2024scaling}.

\myparagraph{Diffusion.}
Given a data distribution $p_{\vx}$ and a noise distribution $p_{\bm{\epsilon}}$,
a diffusion process progressively corrupts clean data \( \vx_0 \sim p_{\vx}\) by adding noise \( \bm{\epsilon} \sim p_{\bm{\epsilon}}\) and then reverses this corruption to recover the original data~\citep{song2019generative,ho2020denoising}, represented as:
\begin{equation}
\label{eq:transform}
\vx_t = \alpha_t \cdot \vx_0 + \sigma_t \cdot \bm{\epsilon},
\end{equation}
where \( t \in [0, \text{T}] \) and \( \bm{\epsilon} \) is drawn from a standard Gaussian distribution, \(p_{\bm{\epsilon}} = \mathcal{N}(0, I) \).
The functions \( \alpha_t \) and \( \sigma_t \) govern the trajectory between clean data and noise, affecting both training and sampling.
The basic parameterization in \citet{ho2020denoising} defines \( \sigma_t = \sqrt{1 - \alpha_t^2} \) with \(\alpha_t = \left( \prod_{s=0}^{t} (1 - \beta_s) \right)^{\frac{1}{2}} \) for discrete timesteps.
The diffusion coefficients \( \beta_t \) are linearly interpolated values between \( \beta_{0} \) and \( \beta_{T-1} \) as \(\beta_t = \beta_0 + \frac{t}{T-1} (\beta_{T-1} - \beta_0)\), with start and end values are set empirically.

The forward and reverse diffusion processes are described by the following factorizations:
\begin{equation}
\begin{gathered}
    q(\vx_{\Delta t:\text{T}} | \vx_0) = \prod_{i=1}^\text{T} q(\vx_{i \cdot \Delta t} | \vx_{(i - 1) \cdot \Delta t})\\
    \textrm{and} \;\; p(\vx_{0:\text{T}}) = p(\vx_\text{T}) \prod_{i=1}^\text{T} p(\vx_{(i - 1) \cdot \Delta t} | \vx_{i \cdot \Delta t}),
\label{eq:fwd}
\end{gathered}
\end{equation}
where the forward process \( q(\vx_{\Delta t:\text{T}} | \vx_0) \) transitions clean data \( \vx_0 \) to noise \( \vx_\text{T} = \bm{\epsilon} \), while the reverse process \( p(\vx_{0:\text{T}}) \) recovers clean data from noise.
$\Delta t$ denotes the time step interval or step size.

During training, the model learns the score function \( \nabla \log p_t(\vx) \propto - \frac{\epsilon}{\sigma_t} \), which represents gradient pointing toward the data distribution along the noise-to-data trajectory.
In practice, the model \( s_\Theta(\vx_t, t) \) is optimized by minimizing the score-matching objective:
\begin{equation}
\label{eq:diffusion}
\gL_\text{score} = \min_\Theta \mathbb{E}_{t \sim \pi(t), \epsilon \sim \mathcal{N}(0, I)} \left[ w_t \|\sigma_t s_\Theta(\vx_t, t) + \bm{\epsilon} \|^2 \right],
\end{equation}
where \( \pi(t) \) defines the time-step sampling distribution and \( w_t \) is a time-dependent weight.
These elements together influence which time steps or noise levels are prioritized during training.
Conceptually, the diffusion model learns the tangent of the trajectory at each point along the path.
During sampling, it progressively recovers clean data from noise based on its predictions.

\myparagraph{Rectified flow} provides a specific parametrization of \( \alpha_t \) and \( \sigma_t \) such that the trajectory between data and noise follows a ``straight'' path~\citep{liu2023flow,albergo2022building}. This trajectory is represented as:
\begin{equation}
\label{eq:rf_traj}
\vx_t = (1 - t) \cdot \vx_0 + t \cdot \bm{\epsilon},
\end{equation}
where \( t \in [0, 1] \). 
In this formulation, the gradient along the trajectory,  \( \bm{\epsilon} - \vx_0 \), is deterministic, often referred to as the velocity.
The model \( v_\Theta (\vx_t, t) \) is parameterized to predict velocity by minimizing:
\begin{equation}
\label{eq:rf}
    \min_\Theta \E_{t \sim \pi(t), \bm{\epsilon} \sim \mathcal{N}(0, I)} \left[ \|v_\Theta (\vx_t, t) - (\bm{\epsilon} - \vx) \|^2 \right].
\end{equation}
We note that this objective is equivalent to a score matching form (\Eqref{eq:diffusion}), with the weight \(w_t = (\frac{1}{1-t})^2 \).
This equivalence highlights that alternative model parameterizations reduce to a standard denoising objective, where the primary difference lies in the time-dependent weighting functions and the corresponding optimization trajectory~\citep{kingma2024understanding}.

During sampling, the model follows a simple probability flow ODE:
\begin{equation}
\label{eq:ode}
    \rd \vx_t = v_\Theta(\vx_t, t) \cdot \rd t. 
\end{equation}
Although a perfect straight path could theoretically be solved in a single step, the independent coupling between data and noise often results in curved trajectories, necessitating multiple steps to generate high-quality samples~\citep{liu2023flow,lee2024improving}.
In practice, we iteratively apply the standard Euler solver~\citep{euler1845institutionum} to sample data from noise.

 \section{Method}
\label{sec:method}

\begin{figure*}[t]
\vskip 0.02in
\begin{center}
\includegraphics[width=\linewidth]{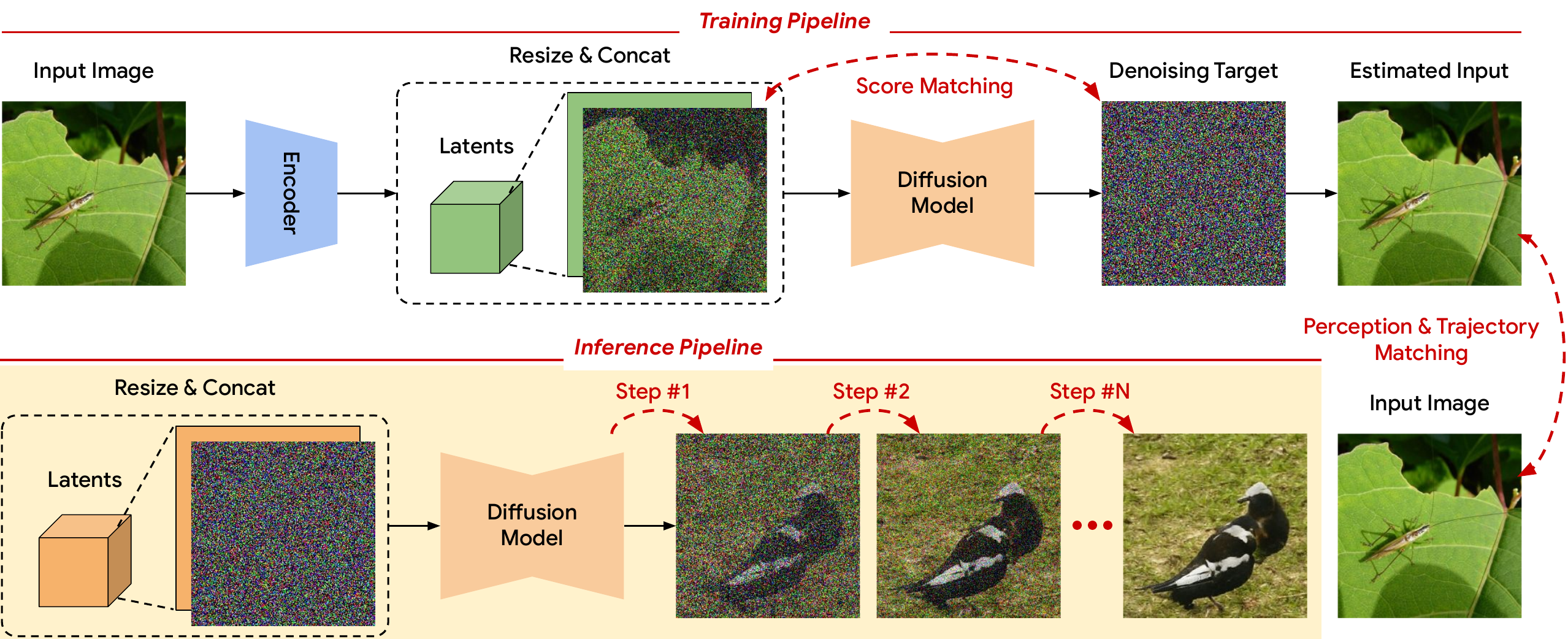}
\vskip -0.06in
\caption{
\textbf{An overview of \OURS}.
We frame visual decoding as an iterative denoising problem by replacing the autoencoder decoder with a diffusion model, optimized using a combination of score, perception, and trajectory matching losses.
During inference, images are reconstructed (or generated) from encoded (or sampled) latents through an iterative denoising process. The number of sampling steps $N$ can be flexibly adjusted within small NFE regimes (from 1 to 3).
We empirically confirm that \OURS significantly outperforms the standard VAE schema, even with just a few steps.
}
\label{fig:pipeline}
\end{center}
\vskip -0.1in
\end{figure*}

We introduce \OURS, with an overview provided in \cref{fig:pipeline}.
The core idea is to replace single-step, deterministic decoding with an iterative, stochastic denoising process. 
By reframing autoencoding as a conditional denoising problem, we anticipate two key improvements: (1) more effective generation of latent representations, allowing the downstream latent diffusion model to learn more efficiently, and (2) enhanced decoding quality due to the iterative and stochastic nature of the diffusion process.

We systematically explore the design space of model architecture, objectives, and diffusion training configurations, including noise and time scheduling.
While this work primarily focuses on generating continuous latents for latent diffusion models, the concept of iterative decoding could also be extended to discrete tokens, which we leave for future exploration.

\subsection{Modeling}

\OURS retains the encoder $\gE$ while enhancing the decoder $\gG$ by incorporating a diffusion model, transforming the standard decoding process into an iterative denoising task. 

\myparagraph{Conditional denoising.}
 Specifically, the input $\vx \sim p_{\vx}$ is encoded by the encoder as $\vz = \gE(\vx)$, and this encoding serves as a condition to guide the subsequent denoising process.
This reformulates the reverse process in \Eqref{eq:fwd} into a conditional form~\citep{nichol2021improved}:
\begin{equation}
\label{eq:aefwd}
    p(\vx_{0:\text{T}}|\vz) = p(\vx_\text{T}) \prod_{i=1}^\text{T} p(\vx_{(i - 1) \cdot \Delta t} | \vx_{i \cdot \Delta t}, \vz),
\end{equation}
where the denoising process from the noise $\vx_\text{T}=\bm{\epsilon}$ to the input $\vx_0=\vx$, is additionally conditioned on $\vz$ over time. Here, the decoder is no longer deterministic, as the process starts from random noise.
For a more detailed discussion on this autoencoding formulation, we refer readers to \cref{sec:discussion}.

\myparagraph{Architecture and conditioning.}
We adopt the standard U-Net architecture from \citet{dhariwal2021diffusion} for our diffusion decoder $\gG$, while also exploring Transformer-based models~\citep{peebles2023scalable}.
For conditional denoising, we concatenate the conditioning signal with the input channel-wise, following the approach of diffusion-based super-resolution models~\citep{ho2022cascaded,saharia2022image}. Specifically, low-resolution latents are upsampled using nearest-neighbor interpolation to match the resolution of $\vx_t$, then concatenated along the channel dimension. 
In \cref{app:exp:additional}, although we experimented with conditioning via AdaGN~\citep{nichol2021improved}, it did not yield significant improvement and introduced additional overhead, so we adopt channel concatenation.

\subsection{Objectives}

We adopt the standard autoencoding objective from \Eqref{eq:vqgan} to train \OURS, with a key modification: replacing the reconstruction loss \( \gL_\text{rec} \) used for the standard decoder with the score-matching loss \( \gL_\text{score} \) for training the diffusion decoder.
Additionally, we introduce a strategy to adjust the perceptual \( \gL_\text{LPIPS} \) and adversarial \( \gL_\text{adv} \) losses to better align with the diffusion decoder training.

\myparagraph{Velocity prediction.}
We adopt the rectified flow parameterization, utilizing a linear optimization trajectory between data and noise, combined with velocity-matching objective (\Eqref{eq:rf}). We inject the encoder output $\vz$ into the objective by replacing $v_\Theta (\vx_t, t)$ with $\gG (\vx_t, t, \vz)$.

\myparagraph{Perceptual matching.} The LPIPS loss~\citep{zhang2018unreasonable} minimizes the perceptual distance between the reconstructions and real images using pre-trained models, typically VGG network~\citep{esser2021taming,yu2023magvit,yu2022vector}. 
We apply this feature-matching objective to train \OURS. 
However, unlike traditional autoencoders, \OURS predicts velocity instead of directly reconstructing the image during training, making it infeasible to compute the LPIPS loss directly between the prediction and the target image.
To address this, we leverage the simple reversing step from \Eqref{eq:rf} to estimate $\vx_0$ from the prediction and $\vx_t$ as follows:
\begin{equation}
\label{eq:rf_est}
    \hat{\vx}_0^t = \vx_t - t \cdot \gG (\vx_t, t, \vz),
\end{equation}
where $\hat{\vx}_0^t$ represents the reconstructed image estimated by the model at time $t$.
We then compute the LPIPS loss between $\hat{\vx}_0^t$ and the target real image $\vx$.

\myparagraph{Denoising trajectory matching.}
The adversarial loss encourages photorealistic outputs by comparing the reconstructions to real images.
We modify this to better align with a diffusion decoder.
Specifically, our approach adapts the standard adversarial loss to enforce trajectory consistency rather than solely on realism.
In practice, we achieve this by minimizing the following divergence, $\gD_\text{adv}$:
\begin{equation}
\label{eq:div}
\min_\Theta \E_{t \sim p_t}\left[ \gD_\text{adv} \left( q(\vx_0|\vx_t) || p_\Theta(\hat{\vx}_0^t|\vx_t) \right) \right],
\end{equation}
where $\gD_\text{adv}$ is a probability distance metric~\citep{goodfellow2014generative,arjovsky2017wasserstein}, and we adopt the basic non-saturating GAN~\citep{goodfellow2014generative}.

For adversarial training, we design a time-dependent discriminator that takes time as input using AdaGN approach~\citep{dhariwal2021diffusion}. To simulate the trajectory, we concatenate $\vx_0$ and $\vx_t$ along the channel dimension.
The generator parameterized by \( \Theta \), and the discriminator, parameterized by \( \Phi \), are then optimized through a minimax game as:
\begin{multline}
\label{eq:adv}
\min_{\Theta} \max_{\Phi} \gL_\text{adv} = \E_{q(\vx_0|\vx_t)}\left[ \log\gD_\Phi(\vx_0, \vx_t, t) \right]\\ + \E_{p_\Theta(\hat{\vx}_0^t|\vx_t)}\left[ \log\left(1 - \gD_\Phi(\hat{\vx}_0^t, \vx_t, t)\right) \right],
\end{multline}
where fake trajectories $p_\Theta(\hat{\vx}_0^t|\vx_t)$ are contrasted with real trajectories $q(\vx_0|\vx_t)$.
To further stabilize training, we apply the $R_1$ gradient penalty to the discriminator parameters~\citep{mescheder2018training}.
In \cref{app:exp:additional}, we explore alternative matching approaches, including the standard adversarial method of comparing individual reconstructions \( \hat{\vx}_0^t \) with real images \( \vx_0 \), matching the trajectory steps $\vx_{t} \rightarrow \vx_{t - \Delta t}$~\citep{xiao2021tackling,wang2024phased}, and our start-to-end trajectory matching $\vx_{t} \rightarrow \vx_0$, with the latter showing the best performance.

\textbf{Final training objective} combines $\gL_\text{score}$, $\gL_\text{LPIPS}$, and $\gL_\text{adv}$, with empirically adjusted weights (see \cref{app:exp:imp}).

Note that applying LPIPS and adversarial losses on the estimated one-step sample could lead to potential objective bias. However, we would like to emphasize that \OURS differs significantly from traditional diffusion models in that its diffusion decoder is conditioned on encoded latents $\vz$. This conditioning provides a strong prior about the input image to reconstruct, resulting in a more accurate estimated one-step sample than in typical diffusion scenarios. Therefore, we believe the potential for objective bias is considerably reduced in \OURS. Fine-tuning the diffusion decoder with frozen $\vz$ like~\citet{sargent2025flow} could be a promising avenue for further improvement, which we will explore in our future work.

\subsection{Noise and time scheduling}

\myparagraph{Noise scheduling.} In diffusion models, noise scheduling involves progressively adding noise to the data over time by defining specific functions for \( \alpha_t \) and \( \sigma_t \) in \Eqref{eq:transform}.
This process is crucial as it determines the signal-to-noise ratio, \( \lambda_t = \frac{\alpha_t^2}{\sigma_t^2} \), which directly influences training dynamics.
Noise scheduling can also be adjusted by scaling the intermediate states \( \vx_t \) with a constant factor $\gamma \in (0, 1]$, which shifts the signal-to-noise ratio downward. 
This makes training more challenging over time while preserving the shape of the trajectory~\citep{chen2023importance}.

In this work, we define \( \alpha_t \) and \( \sigma_t \) according to rectified flow formulation, while also scaling \( \vx_t \) by $\gamma$, with the value chosen empirically. However, when $\gamma \neq 1$, the variance of $\vx_t$ changes, which can degrade performance~\citep{karras2022elucidating}. 
To address this, we normalize the denoising input $\vx_t$ by its variance after scaling, ensuring it preserves unit variance over time~\citep{chen2023importance}.

\myparagraph{Time scheduling.}
Another important aspect in diffusion models is time scheduling for both training and sampling, controlled by \( \pi(t) \) during training and \( \Delta t \) during sampling, as outlined in \Eqref{eq:fwd} and \Eqref{eq:diffusion}.
A common choice for \( \pi(t) \) is the uniform distribution $\gU(0,T)$, which applies equal weight to each time step during training. 
Similarly, uniform time steps \( \Delta t = \frac{1}{T} \) are typically used for sampling.
However, to improve model performance on more challenging time steps and focus on noisy regimes during sampling, the time scheduling strategy should be adjusted accordingly.

In this work, we sample $t$ from a logit-normal distribution~\citep{atchison1980logistic}, which emphasizes intermediate timesteps~\citep{esser2024scaling}.
During sampling, we apply a reversed logarithm mapping $\rho_\text{log}$ defined as:
\begin{equation}
    \rho_\text{log}(t; m, n) = \frac{\log(m) - \log\left(t \cdot (m - n) + n\right)}{\log(m) - \log(n)},
\end{equation}
where we set $m = 1$ and $n = 100$, resulting in denser sampling steps early in the inference process.

 \section{Experiments}

We evaluate the effectiveness of \OURS on image reconstruction and generation tasks using the ImageNet~\citep{deng2009imagenet}. 
The VAE formulation by \citet{esser2021taming} serves as a strong baseline due to its widespread use in modern image generative models~\citep{rombach2022high,peebles2023scalable,esser2024scaling}. 
We perform controlled experiments to compare reconstruction and generation quality by varying model scale, latent dimension, downsampling rates, and input resolution.

\myparagraph{Model configurations.} 
We use the encoder and discriminator architectures from VQGAN~\citep{esser2021taming} and keep consistent across all models. 
The decoder design follows BigGAN~\citep{brock2019large} for VAE and from ADM~\citep{dhariwal2021diffusion} for \OURS. 
Additionally, we experiment with the DiT architecture~\citep{peebles2023scalable} for \OURS.
To evaluate model scaling, we test five decoder variants: base (B), medium (M), large (L), extra-large (XL), and huge (H), by adjusting width and depth accordingly.
Further model specifications are provided in \cref{app:exp:spec}.

We experiment with the following two encoder configurations. 
\textbf{\OURS-lite}: a light-weight version with 6M parameters, a downsampling rate of 16, and 8 latent channels; 
\textbf{\OURS-SD}: a standard version based on Stable Diffusion with 34M parameters, a downsampling rate of 8, and 4 latent channels. 
\OURS-lite is intentionally designed as a more challenging setup and serves as the primary focus of analysis in the paper.
For this configuration, we further explore downsampling rates of 4, 8, and 32, as well as latent dimensions of 4, 16, and 32 channels.
Both VAE and \OURS are trained to reconstruct $128 \times 128$ images under these controlled conditions.
Additionally, we validate our method in the standard setup of \OURS-SD, where we compare it against state-of-the-art VAEs.

\myparagraph{Evaluation.}
We evaluate the autoencoder on both reconstruction and generation quality using Fréchet Inception Distance (FID) \citep{heusel2017gans} as the primary metric, and we also report PSNR and SSIM as secondary metrics.
For reconstruction quality (rFID), FID is computed at both training and higher resolutions to assess generalization across resolutions.
For generation quality (FID), we generate latents from the trained autoencoders and use them to train the DiT-XL/2 latent generative model~\citep{peebles2023scalable}. 
This latent model remains fixed across all generation experiments, meaning improved autoencoder latents directly enhance generation quality.

\subsection{Reconstruction quality}
\label{sec:exp:reconstruction}

\begin{figure*}[t]
\vskip 0.1in
\begin{center}
\includegraphics[height=11.7em]{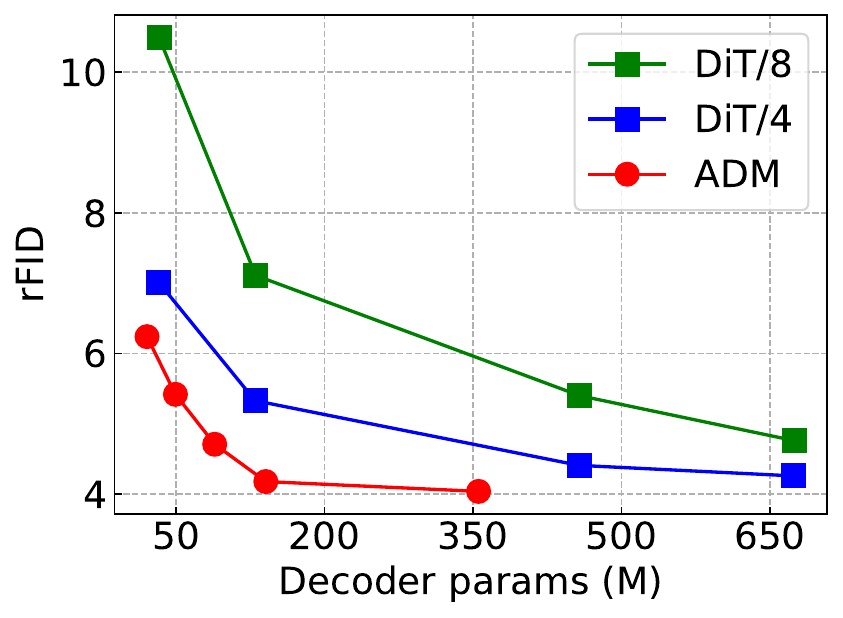}
\includegraphics[height=11.7em]{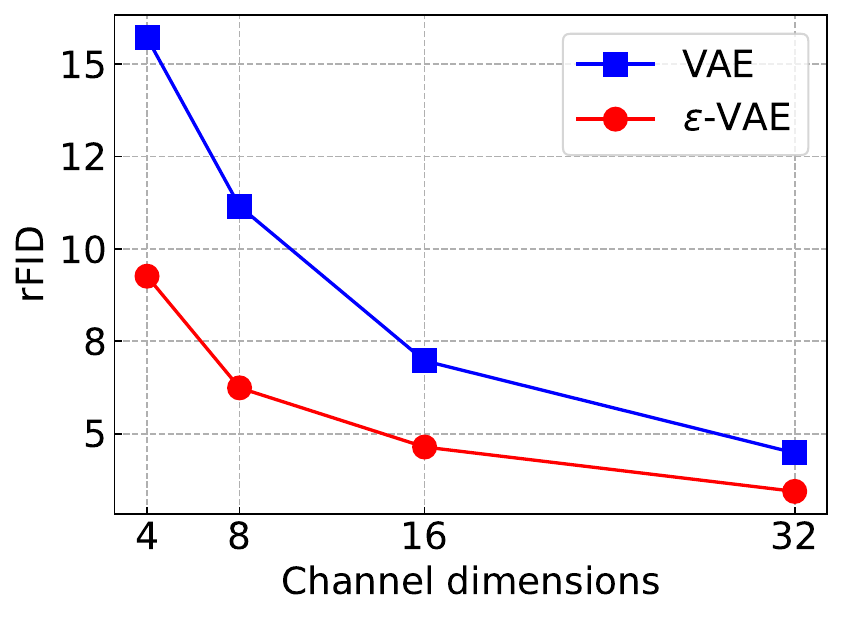}
\includegraphics[height=11.7em]{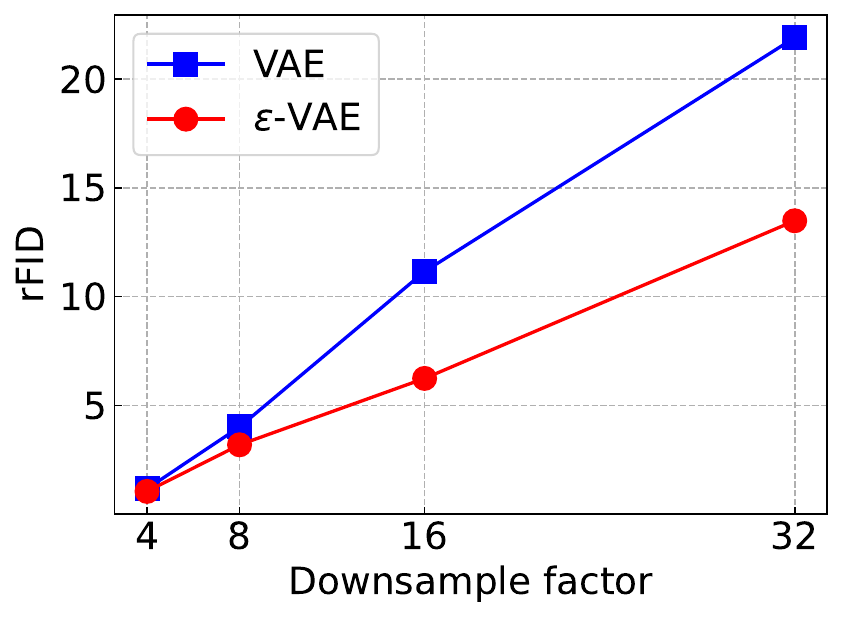}
\vskip -0.1in
\caption{
\textbf{Architecture and compression analysis.} The \OURS decoder uses either a UNet-based ADM or Transformer-based DiT (left). \OURS and VAE are evaluated under different compression rates by varying latent channel dimensions (middle) or encoder downsampling factors (right). \textit{We follow the \OURS-lite configuration in these experiments.} 
}
\label{fig:reconstruct}
\end{center}
\vskip -0.1in
\end{figure*}

\myparagraph{Decoder architecture.}
We explore two major architectural designs:
the UNet-based architecture from ADM~\citep{dhariwal2021diffusion} and the Transformer-based DiT~\citep{peebles2023scalable}.
We compare various model sizes--ADM:\{B, M, L, XL, H\} and DiT:\{S, B, L, XL\} with patch sizes of \{4, 8\}.
The results are summarized in \cref{fig:reconstruct}~(left).
ADM consistently outperforms DiT across the board.
While we observe rFID improvements in DiT when increasing the number of tokens by reducing patch size, this comes with significant computational overhead.
The overall result aligns with the original design intentions: ADM for pixel-level generation and DiT for latent-level generation.
For the following experiments, we use the ADM architecture for our diffusion decoder.

\myparagraph{Compression rate.} 
Compression can be achieved by adjusting either the channel dimensions of the latents or the downsampling factor of the encoder. 
In \cref{fig:reconstruct} (middle and right), we compare VAE and \OURS across these two aspects.
The results show that \OURS consistently outperforms VAE in terms of rFID, particularly as the compression ratio increases. Specifically, as shown on the middle graph, \OURS achieves lower rFIDs than VAE across all channel dimensions, with a notable gap at lower dimensions (4 and 8). On the right graph, \OURS maintains lower rFIDs than VAE even as the downsampling factor increases, with the gap widening significantly at larger factors (16 and 32). Furthermore, \OURS delivers comparable or superior rFIDs even when the compression ratio is doubled, demonstrating its robustness and effectiveness in high-compression scenarios.

\begin{table}[t]
\caption{\textbf{ImageNet reconstruction results (rFID) at different resolutions using VAEs trained at $128 \times 128$.} $\dag$ denotes training at $128 \times 128$ followed by fine-tuning at a higher resolution.
}
\label{tbl:generalization}
\vskip 0.1in
\begin{center}
\resizebox{0.96\linewidth}{!}{\begin{tabular}{l|ccc|c}
\toprule
\textbf{Resolution (ImageNet)} & $\mathbf{128}$ & $\mathbf{256}$ & $\mathbf{512}$ & $\mathbf{256}^\dag$ \\
\midrule
SD-VAE~\citep{rombach2022high} & 4.54 & 1.21 & 0.91 & 0.86 \\
LiteVAE~\citep{sadat2024litevae} & 4.40 & 0.97 & - & 0.73 \\
\midrule
\OURS-SD (B) & 1.94 & 0.65 & 0.61 & 0.52 \\
\OURS-SD (M) & 1.58 & 0.55 & 0.53 & 0.47 \\
\OURS-SD (L) & 1.47 & 0.52 & 0.41 & 0.45 \\
\OURS-SD (XL) & 1.34 & 0.49 & 0.39 & 0.43 \\
\OURS-SD (H) & 1.00 & 0.44 & 0.35  & 0.38 \\
\bottomrule
\end{tabular}
}
\end{center}
\vskip -0.1in
\end{table}

\myparagraph{Resolution generalization.} 
A notable feature of conventional autoencoders is their capacity to generalize and reconstruct images at higher resolutions during inference~\citep{rombach2022high}. 
To assess this, we conduct inference on images with resolutions of $256 \times 256$ and $512 \times 512$, using \OURS and VAE models trained at $128 \times 128$. 
As shown in \cref{tbl:generalization}, \OURS effectively generalizes to higher resolutions, consistently preserving its performance advantage over other VAEs. Furthermore, we find that fine-tuning models at the target (higher) resolution leads to improvement at it, which is consistent with the observation made by~\citet{sadat2024litevae}. We hence utilize this multi-stage training strategy in the following experiments when the target resolution is larger than $128 \times 128$.

\begin{table*}[t]
\caption{\textbf{Comparisons with state-of-the-art image autoencoders.} All results are computed on $256 \times 256$ ImageNet 50K validation set and COCO-2017 5K validation set. \OURS-SD (M) achieves better reconstruction quality while having similar parameters (49M) in the decoder with other VAEs. Further improvements are obtained after we scale up to \OURS-SD (H) which has 355M decoder parameters.
}
\label{tbl:sd-vae}
\vskip 0.1in
\begin{center}
\resizebox{0.9\linewidth}{!}{\begin{tabular}{c|lcc|ccc|ccc}
\toprule
\multirowcell{2}{\textbf{Downsample}\\ \textbf{factor}} & \multirow{2}{*}{\textbf{Method}} & \multirowcell{2}{\textbf{Discrete}\\ \textbf{latent}} & \multirowcell{2}{\textbf{Latent}\\ \textbf{dim.}} & \multicolumn{3}{c|}{\textbf{ImageNet}} & \multicolumn{3}{c}{\textbf{COCO}} \\
& & & & \textbf{rFID}$\downarrow$ & \textbf{PSNR}$\uparrow$ & \textbf{SSIM}$\uparrow$ & \textbf{rFID}$\downarrow$ & \textbf{PSNR}$\uparrow$ & \textbf{SSIM}$\uparrow$ \\
\midrule
\multirow{6}{*}{$16 \times 16$} & VQGAN~\citep{esser2021taming} & \cmark & 256 & 4.99 & 20.00 & 0.629 & 12.29 & 19.57 & 0.630 \\
& MaskGIT~\citep{chang2022maskgit} & \cmark & 256 & 2.28 & - & - & - & - & - \\
& LlamaGen~\citep{sun2024autoregressive} & \cmark & 8 & 2.19 & 20.79 & 0.675 & 8.11 & 20.42 & 0.678 \\
& SD-VAE~\citep{rombach2022high} & \xmark & 4 & 2.93 & 20.57 & 0.662 & 8.89 & 19.95 & 0.670 \\
& \bluecell \OURS-SD (M) & \bluecell \xmark & \bluecell 4 & \bluecell \underline{1.91} & \bluecell \underline{21.27} & \bluecell \underline{0.693} & \bluecell \underline{6.12} & \bluecell \underline{22.38} & \bluecell \underline{0.718} \\
& \bluecell \OURS-SD (H) & \bluecell \xmark & \bluecell 4 & \bluecell \textbf{1.35} & \bluecell \textbf{22.60} & \bluecell \textbf{0.711} & \bluecell \textbf{4.18} & \bluecell \textbf{24.26} & \bluecell \textbf{0.830} \\
\midrule
\multirow{8}{*}{$8 \times 8$}& VQGAN~\citep{esser2021taming} & \cmark & 4 & 1.19 & 23.38 & 0.762 & 5.89 & 23.08 & 0.771 \\
& ViT-VQGAN~\cite{yu2022vector} & \cmark & 32 & 1.28 & - & - & - & - & - \\
& LlamaGen~\citep{sun2024autoregressive} & \cmark & 8 & 0.59 & 24.45 & 0.813 & 4.19 & 24.20 & 0.822 \\
& SD-VAE~\citep{rombach2022high} & \xmark & 4 & 0.74 & 25.68 & 0.820 & 4.45 & 25.41 & 0.831 \\
& SDXL-VAE~\citep{podell2023sdxl} & \xmark & 4 & 0.68 & 26.04 & 0.834 & 4.07 & 25.76 & 0.845 \\
& LiteVAE~\cite{sadat2024litevae} & \xmark & 4 & 0.87 & 26.02 & 0.740 & - & - & - \\
& \bluecell \OURS-SD (M) & \bluecell \xmark & \bluecell 4 & \bluecell \underline{0.47} & \bluecell \underline{27.65} & \bluecell \underline{0.841} & \bluecell \underline{3.98} & \bluecell \underline{25.88} & \bluecell \underline{0.850} \\
& \bluecell \OURS-SD (H) & \bluecell \xmark & \bluecell 4 & \bluecell \textbf{0.38} & \bluecell \textbf{29.49} & \bluecell \textbf{0.851} & \bluecell \textbf{3.65} & \bluecell \textbf{26.01} & \bluecell \textbf{0.856} \\
\bottomrule
\end{tabular}
}
\end{center}
\vskip -0.1in
\end{table*}

\myparagraph{Comparisons to state-of-the-art VAEs.} We provide image reconstruction results under the same configuration as VAEs in Stable Diffusion (SD-VAE): an encoder with 34M parameters and a channel dimension of 4 for $256 \times 256$ image reconstruction. We evaluate rFID, PSNR and SSIM on the full validation sets of ImageNet and COCO-2017~\citep{lin2014microsoft}, with the results summarized in \cref{tbl:sd-vae}. Our finds reveal that \OURS outperforms state-of-the-art VAEs when the decoder sizes are comparable, and its performance can be further improved by scaling up the decoder. This demonstrates the strong model scalability of our framework.

\begin{table}[t]
\caption{\textbf{Image reconstruction results of one-step \OURS and SD-VAE on ImageNet $256 \times 256$.} A downsampling factor of $8 \times 8$ is used for comparison. We include two variants of our model in the results: \OURS-SD (B), which has a similar inference speed to SD-VAE, and \OURS-SD (M), which matches SD-VAE in the number of parameters. 
}
\label{tbl:one-step}
\vskip 0.1in
\begin{center}
\resizebox{0.90\linewidth}{!}{\begin{tabular}{l|ccc}
\toprule
\textbf{Method} & \textbf{rFID}$\downarrow$ & \textbf{PSNR}$\uparrow$ & \textbf{SSIM}$\uparrow$ \\
\midrule
SD-VAE~\citep{rombach2022high} & 0.74 & 25.68 & 0.820 \\
\bluecell \OURS-SD (B) & \bluecell \underline{0.57} & \bluecell \underline{25.91} & \bluecell \underline{0.826} \\
\bluecell \OURS-SD (M) & \bluecell \textbf{0.51} & \bluecell \textbf{26.45} & \bluecell \textbf{0.830} \\
\bottomrule
\end{tabular}
}
\end{center}
\vskip -0.1in
\end{table}

\myparagraph{One-step \OURS.} Note that the denoising process of \OURS demonstrates promising results even with a single iteration. To show this, we provide a direct comparison between SD-VAE and our one-step \OURS models in \cref{tbl:one-step}. This table presents image reconstruction quality on ImageNet $256 \times 256$ with the $8 \times 8$ downsampling factor. As shown, both \OURS (B) and \OURS (M) outperform SD-VAE across all metrics. These results confirm the effectiveness and efficiency of our one-step models compared to SD-VAE. Consequently, this allows \OURS to be adapted for scenarios with latency-sensitive requirements, \eg, real-time visualization during image generation, by reducing the decoding step to a single pass.

\subsection{Class-conditional image generation}

\begin{table}[t]
\caption{\textbf{Benchmarking class-conditional image generation on ImageNet $256 \times 256$.} We use the DiT-XL/2 architecture~\citep{esser2024scaling} for latent diffusion models, and we do not apply classifier-free guidance~\citep{ho2022classifier}.
}
\label{tbl:cond-ldm}
\vskip 0.1in
\begin{center}
\resizebox{0.98\linewidth}{!}{\begin{tabular}{c|l|cc}
\toprule
\multirowcell{2}{\textbf{Downsample}\\ \textbf{factor}} & \multirow{2}{*}{\textbf{Method}} & \multirowcell{2}{\textbf{Throughput}\\ \textbf{(image/ks)}} & \multirow{2}{*}{\textbf{FID}$\downarrow$} \\
& \\
\midrule
\multirow{3}{*}{$32 \times 32$} & SD-VAE~\citep{rombach2022high} & 3991 & 21.31 \\
& \bluecell \OURS-SD (M) & \bluecell 3865 & \bluecell \underline{15.98} \\
& \bluecell \OURS-SD (H) & \bluecell 3870 & \bluecell \textbf{14.26} \\
\midrule
\multirow{3}{*}{$16 \times 16$} & SD-VAE~\citep{rombach2022high} & 1220 & 14.59 \\
& \bluecell \OURS-SD (M) & \bluecell \:1192 \tikzmark{a1} & \bluecell \underline{10.68} \\
& \bluecell \OURS-SD (H) & \bluecell 1180 & \bluecell \textbf{9.72} \\
\midrule
\multirow{5}{*}{$8 \times 8$} & Asym-VAE~\citep{zhu2023designing} & 502 & 10.85 \\
& Omni-VAE~\citep{wang2024omnitokenizer} & 480 & 12.25 \\
& SD-VAE~\citep{rombach2022high} & \:522 \tikzmark{t1} & 11.63 \\
& \bluecell \OURS-SD (M) & \bluecell 491 & \bluecell \underline{9.39} \\
& \bluecell \OURS-SD (H) & \bluecell 477 & \bluecell \textbf{8.85} \\
\bottomrule
\end{tabular}
\begin{tikzpicture}[overlay, remember picture, shorten >=.5pt, shorten <=.5pt, transform canvas={yshift=.25\baselineskip}]
\draw [-{Stealth[length=2.4mm, width=1.6mm]}, red] ({pic cs:a1}) [bend left] to node [below right] (t1) {\hspace{-2pt}\small \textbf{2.3$\times$}} ({pic cs:t1});
\end{tikzpicture}
}
\end{center}
\vskip -0.1in
\end{table}

We now evaluate the generative performance of \OURS when combined with latent diffusion models~\citep{rombach2022high}. We perform standard class-conditional image generation tasks using the DiT-XL/2 model as our latent generative model~\citep{peebles2023scalable}. Further details on the training setup are provided in \cref{app:exp:ldm}. \cref{tbl:cond-ldm} presents the image generation results of \OURS and other competing VAEs at resolutions of $256 \times 256$.
The results show that \OURS consistently outperforms other VAEs across different dowmsampling factors. In addition, we emphasize that \OURS achieves favorable generation quality while using only 25\% of the token length typically required by SD-VAE. This token length reduction significantly accelerates latent diffusion model generation, leading to 2.3$\times$ higher inference throughput while maintaining competitive generation quality. These results confirm that the performance gains from the reconstruction task successfully transfer to the generation task, further validating the effectiveness of \OURS.

More importantly, \OURS-SD achieves around 25\% improvement in generation quality over SD-VAE at the $32 \times 32$ downsampling factor, alongside a 3.2$\times$ inference speedup than SD-VAE at the $16 \times 16$ downsampling factor with comparable FID. We observed similar training speedups for latent diffusion models utilizing \OURS at this higher downsampling rate. These gains are more pronounced than those observed when increasing the downsampling factor from $8 \times 8$ to $16 \times 16$. These findings strongly suggest that the benefits of \OURS and latent diffusion pipeline could be amplified with higher downsampling factors.

An additional advantage of scaling the autoencoder over the latent model lies in computational efficiency. Recent trends show latent diffusion models increasingly adopt Transformer architectures~\citep{peebles2023scalable}, where self-attention scales quadratically with input resolution. In contrast, our convolution-based UNet decoder offers more favorable linear scaling. As models grow, shifting complexity to the autoencoder helps reduce the burden on the latent model, leading to a more efficient overall system.

\begin{table*}[t]
\caption{\textbf{Ablation study on key design choices for the \OURS diffusion decoder.} A systematic evaluation of the proposed architecture ($\star$), objectives ($\dag$), and noise \& time scheduling ($\S$).
Each row progressively modifies or builds upon the baseline decoder, showing improvements in performance. \textit{The results are computed under the \OURS-lite configuration.}
}
\label{tbl:ablation}
\vskip 0.1in
\begin{center}
\resizebox{0.64\textwidth}{!}{\begin{tabular}{l|cc}
\toprule
\textbf{Ablation} & \textbf{NFE}$\downarrow$ & \textbf{rFID}$\downarrow$ \\
\midrule
\textit{Baseline:} DDPM-based diffusion decoder & 1,000 & 28.22 \\
$^\dag$~\textit{(a)}~Diffusion $\rightarrow$ Rectified flow parameterization & 100 & 24.11 \\
$^\S$~\textit{(b)}~Uniform $\rightarrow$ Logit-normal time step sampling during training & 50 & 23.44 \\
$^\star$~\textit{(c)}~DDPM UNet $\rightarrow$ ADM UNet & 50 & 22.04 \\
$^\dag$~\textit{(d)}~Perceptual matching on $\hat{\vx}_0^t$ and $\vx_0$ & 10 & 11.76 \\
$^\dag$~\textit{(e)}~Adversarial denoising trajectory matching on $(\hat{\vx}_0^t, \vx_t)$ and $(\vx_0, \vx_t)$ & 5 & 8.24 \\
$^\S$~\textit{(f)}~Scale $\vx_t$ by $\gamma = 0.6$ & 5 & 7.08 \\
$^\S$~\textit{(g)}~Uniform $\rightarrow$ Reversed logarithm time spacing during inference & 3 & 6.24 \\
\bottomrule
\end{tabular}
}
\end{center}
\vskip -0.1in
\end{table*}

\begin{figure*}[t]
\vskip 0.1in
\begin{center}
\includegraphics[height=11.7em]{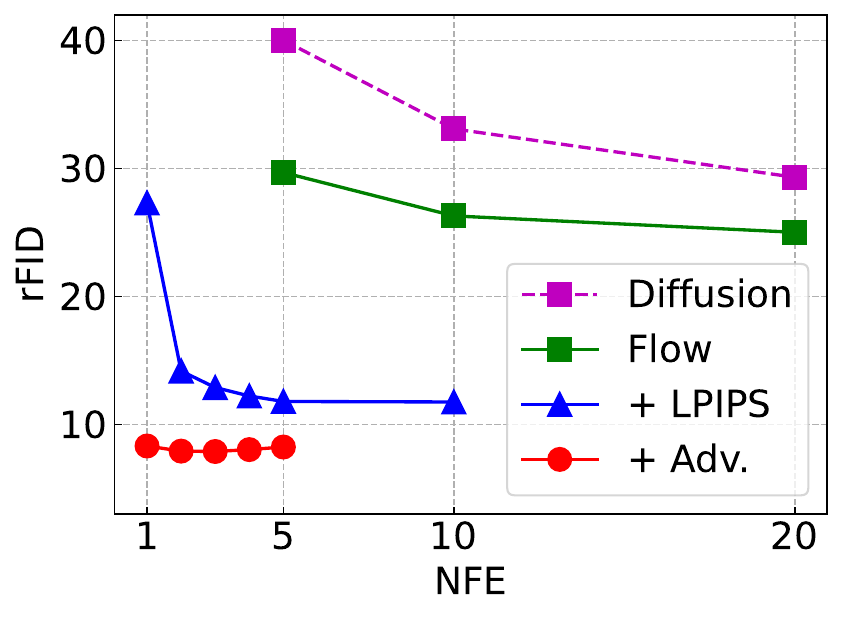}
\includegraphics[height=11.7em]{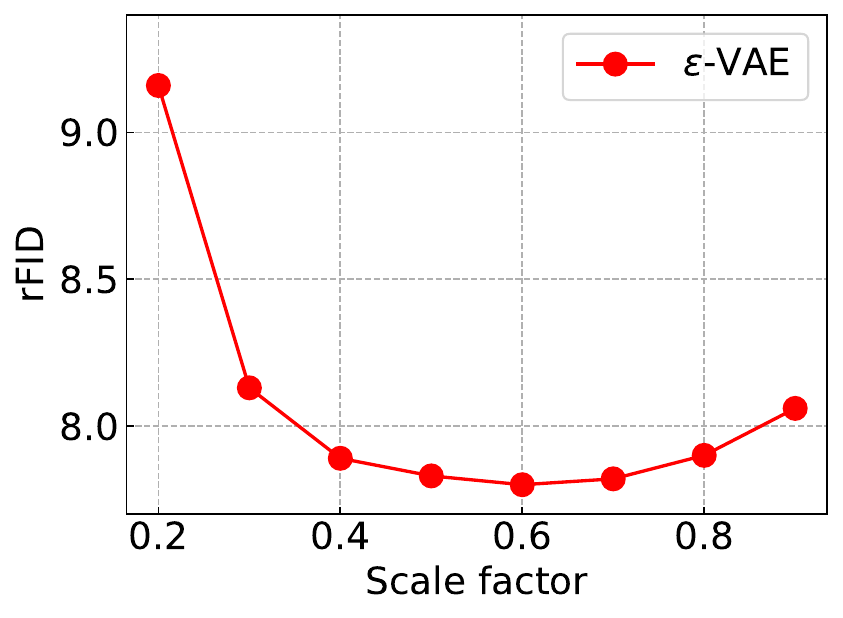}
\includegraphics[height=11.7em]{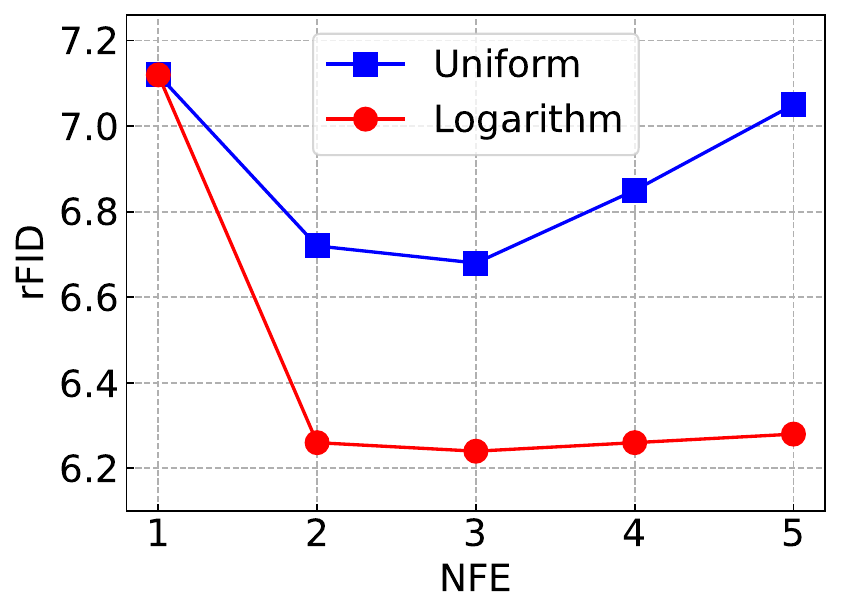}
\vskip -0.1in
\caption{
\textbf{Impact of our major diffusion decoder designs.} 
Improved training objectives, particularly perceptual matching loss and adversarial denoising trajectory matching loss, significantly contribute to better rFID scores and NFE (left).
Effective noise scheduling by modulating the scaling factor $\gamma$ further enhances rFID, with an optimum value of 0.6 in our experiments (middle).
Lastly, adjusting time step spacing during inference ensures stable sampling in low NFE regimes (right). \textit{We report results under the \OURS-lite configuration.}
}
\label{fig:ablation}
\end{center}
\vskip -0.1in
\end{figure*}

\subsection{Ablation studies}
We conduct a component-wise analysis to validate our key design choices, focusing on three critical aspects: architecture, objectives, and noise \& time scheduling.
We evaluate the reconstruction quality (rFID) and sampling efficiency (NFE). The results are summarized in \cref{tbl:ablation}.

\myparagraph{Baseline.} 
Our evaluation begins with a baseline model: an autoencoder with a diffusion decoder, trained solely using the score matching objective.
This baseline follows the vanilla diffusion setup from \citet{ho2020denoising}, including their UNet architecture, parameterization, and training configurations, while extending to a conditional form as described in \Eqref{eq:aefwd}.
Building on this baseline, we progressively introduce updates and evaluate the impact of our proposed method.

\myparagraph{Impact of proposals.}
In \textbf{(a)}, transitioning from standard diffusion to rectified flow~\citep{liu2023flow} straightens the optimization path, resulting in significant gains in rFID and NFE. 
In \textbf{(b)}, adopting a logit-normal time step distribution optimizes rectified flow training~\citep{esser2024scaling}, further improving both rFID and NFE.
In \textbf{(c)}, updates to the UNet architecture~\citep{nichol2021improved} contribute to enhanced rFID scores.
In \textbf{(d)}, LPIPS loss is applied to match reconstructions \( \hat{\vx}_0^t \) with real images \( \vx_0 \). In \textbf{(e)}, adversarial trajectory matching loss aligns $(\hat{\vx}_0^t, \vx_t)$ with $(\vx_0, \vx_t)$, the target transition in rectified flow.
Both objectives improve model understanding of the underlying optimization trajectory, significantly enhancing rFID scores and NFE.

Up to this point, with the full implementation of \Eqref{eq:vqgan}, we can compare our proposal with the VAE (B) model, which achieves an rFID score of 11.15. 
Our model, with a score of 8.24, already surpasses this baseline.
We further improve performance by optimizing noise and time scheduling within our framework, as described next.

In \textbf{(f)}, scaling \( \vx_t \) reduces the signal-to-noise ratio~\citep{chen2023importance}, presenting challenges for more effective learning during training.
\cref{fig:ablation} (middle) demonstrates that a scaling factor of 0.6 produces the best results.
Finally, in \textbf{(g)}, reversed logarithmic time step spacing during inference allows for denser evaluations in noisier regions.
\cref{fig:ablation} (right) demonstrates that this method provides more stable sampling in the lower NFE regime compared to the original uniform spacing.

In \cref{fig:ablation} (right), reconstruction quality degrades when the number of denoising steps exceeds three.
To enable large step sizes for the reverse process during inference, we introduce the denoising trajectory matching loss to implicitly model the conditional distribution $p(\vx_0|\vx_t)$, shifting the denoising distributions from traditional Gaussian to non-Gaussian multimodal forms~\citep{xiao2021tackling}.
However, the assumptions underlying this approach are most effective when the total number of denoising steps is small. 
This reveals an optimal range of one to three inference steps.
The degradation beyond this range also suggests that uniform step spacing may no longer be ideal. 
Accordingly, we empirically explored alternative sampling strategies and found that a reversed logarithmic schedule yields improved performance, as shown in the figure.

\begin{figure*}[t]
\vskip 0.1in
\begin{center}
\includegraphics[width=\linewidth]{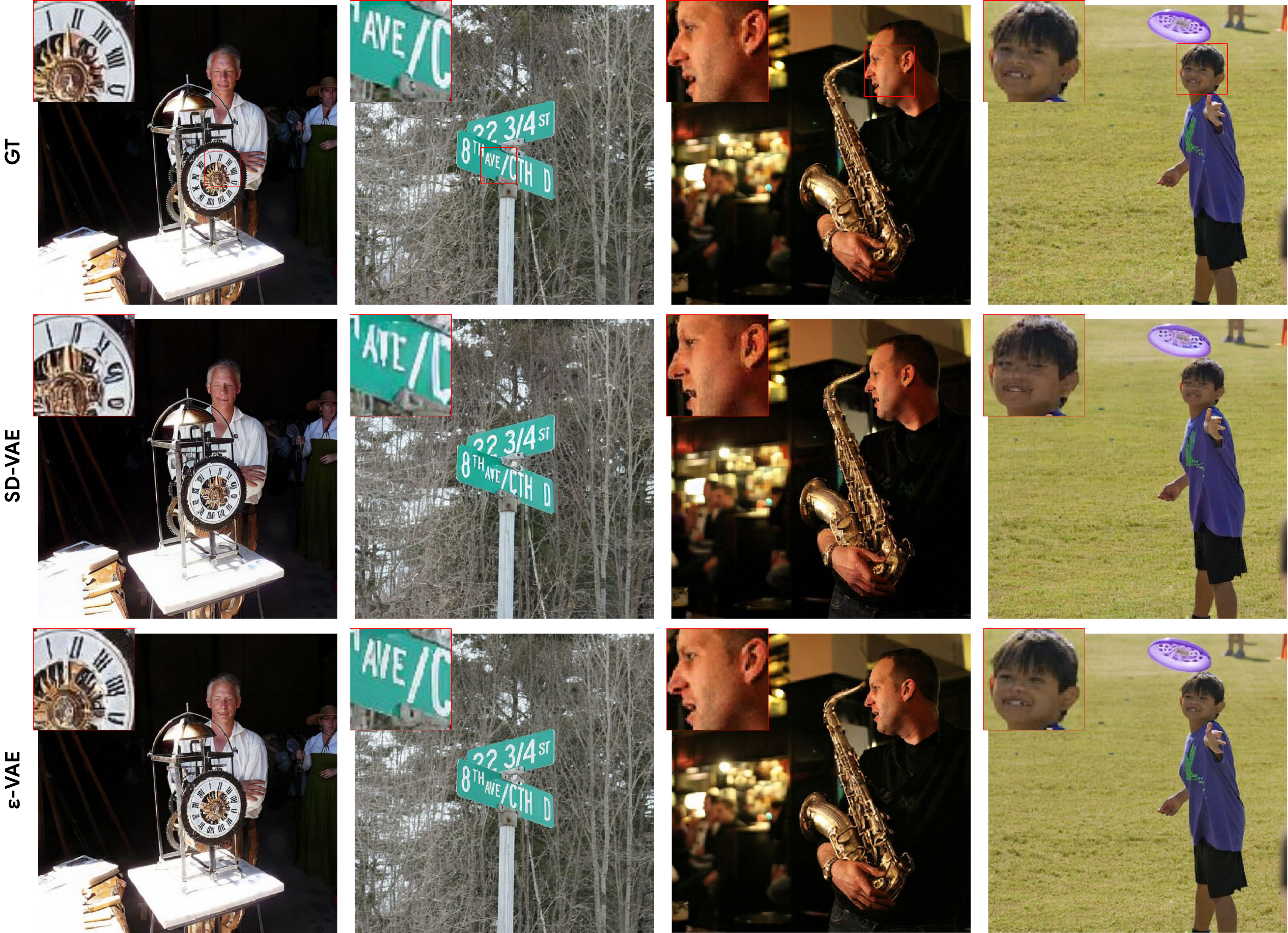}
\vskip -0.06in
\caption{\textbf{Image reconstruction results under the SD-VAE configuration~\citep{rombach2022high} at the resolution of $512 \times 512$.} We find that \OURS produces more accurate visual details than SD-VAE in the highlighted regions with text or human face. 
\textit{Best viewed when zoomed-in and in color}.}
\label{fig:rec512}
\end{center}
\vskip -0.1in
\end{figure*}

\subsection{Visualization}

In addition to the quantitative results, \cref{fig:rec512} shows high-resolution image reconstruction samples produced by SD-VAE~\citep{rombach2022high} and \OURS at the resolution of $512 \times 512$. We observe that reconstructed images generated by \OURS demonstrate a better visual quality than ones of SD-VAE. In particular, \OURS maintains a good visual quality for small text and human face. We provide more visual comparisons in~\cref{app:visual} and throughout discussions on the major properties and advantages of \OURS compared to traditional VAEs in~\cref{sec:discussion}. \section{Conclusion}

We present \OURS, an effective visual tokenizer that introduces a diffusion decoder into standard autoencoders, turning single-step decoding into a multi-step probabilistic process. By exploring key design choices in modeling, objectives, and diffusion training, we demonstrate significant performance improvements. Our approach outperforms traditional autoencoders in both reconstruction and generation quality, particularly in high-compression scenarios. We hope our concept of iterative generation during decoding inspires further advancements in visual autoencoding.

\section*{Acknowledgements}

We would like to thank Xingyi Zhou, Weijun Wang, and Caroline Pantofaru for reviewing the paper and providing feedback. We thank Rui Qian, Xuan Yang, and Mingda Zhang for helpful discussion. We also thank the Google Kauldron team for technical assistance.

\section*{Impact statement}

Our work could lead to improved autoencoding techniques which have the potential to benefit generative modeling across various perspectives, including reducing training time and memory requirements, improving visual qualities, \etc Although our work does not uniquely raise any new ethical challenges, visual generative modeling is a field with several ethical concerns worth acknowledging. For example, there are known issues around bias and fairness, either in the representation of generated images~\citep{menon2020pulse} or
the implicit encoding of stereotypes~\citep{steed2021image}, as well as potential risks in privacy. To ensure that the benefits of this technology are harnessed responsibly, we encourage continued
open discussions in the community around the development
of these new technologies.

\bibliography{icml2025}
\bibliographystyle{icml2025}

\clearpage
\appendix
\begin{figure*}[t]
\vskip 0.1in
\begin{center}
\includegraphics[width=\linewidth]{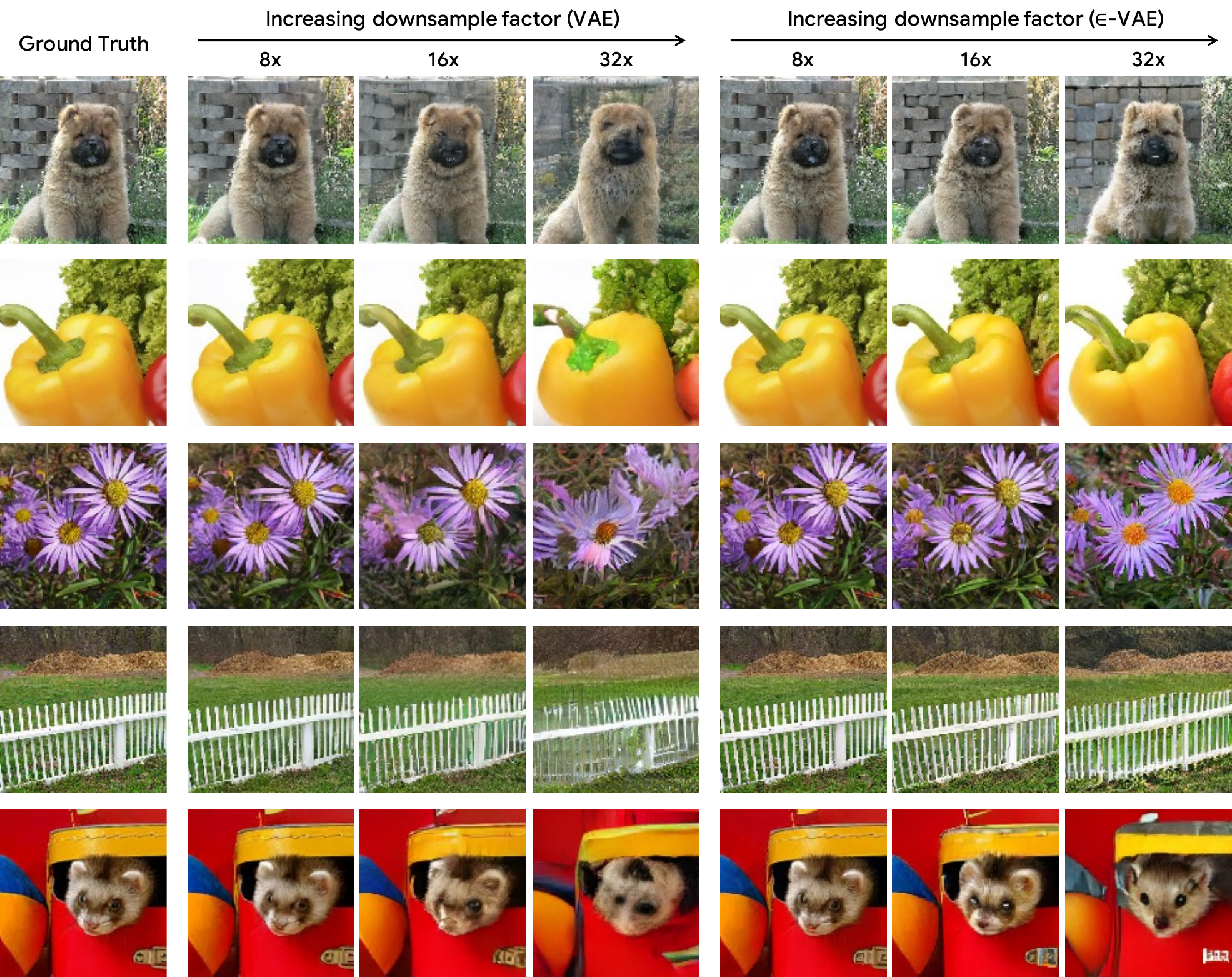}
\vskip -0.1in
\caption{
\textbf{Reconstruction results with varying downsampling ratios.} 
\OURS maintains both high fidelity and perceptual quality, even under extreme downsampling conditions, whereas VAE fails to preserve semantic integrity. \textit{Best viewed when zoomed-in and in color}.
}
\label{fig:viz_down}
\end{center}
\vskip -0.1in
\end{figure*}

\begin{figure*}[t]
\vskip 0.1in
\begin{center}
\includegraphics[width=\linewidth]{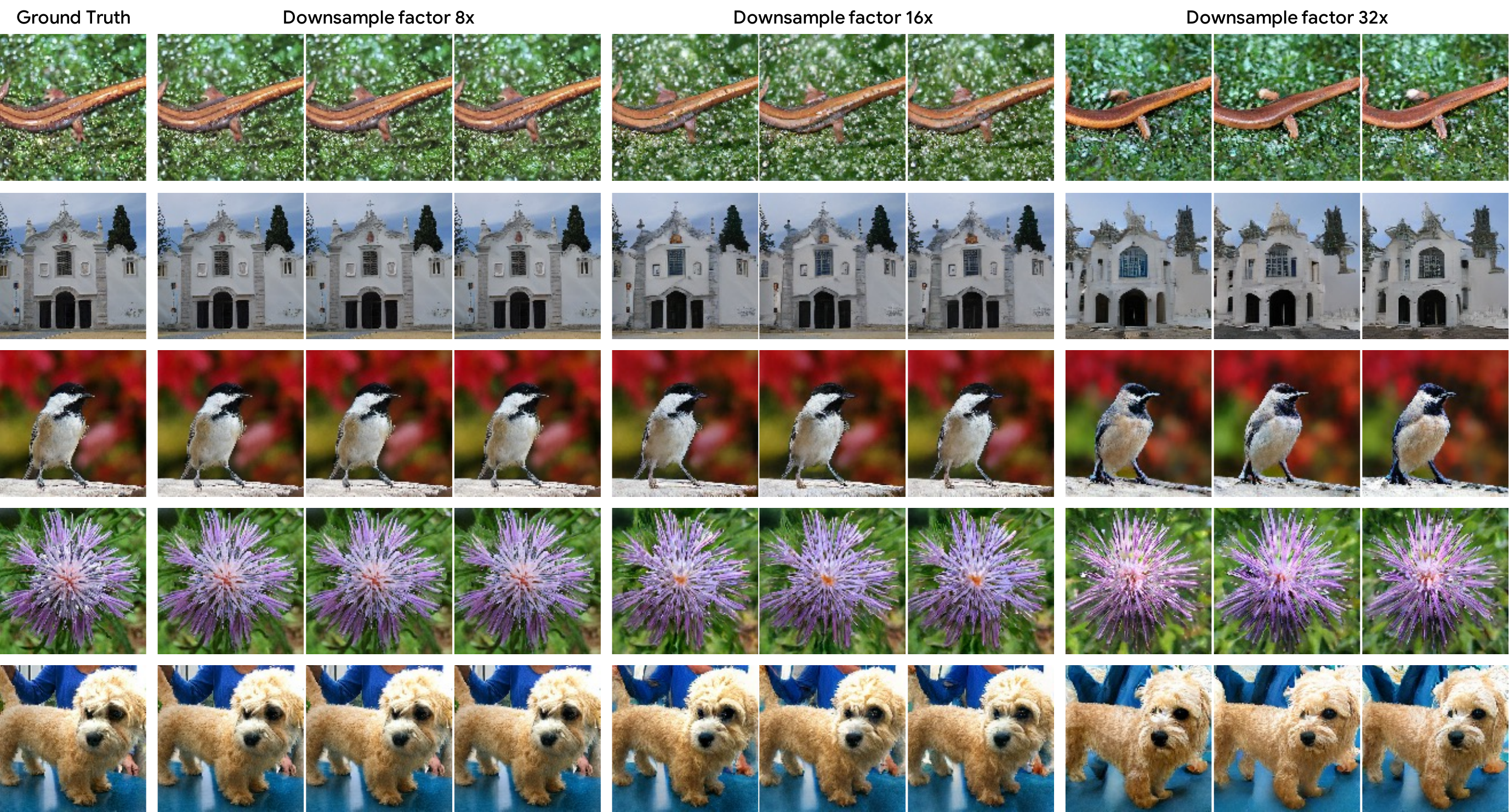}
\vskip -0.1in
\caption{
\textbf{\OURS reconstruction results with varying random seeds and downsampling ratios.}
At lower compression levels, the reconstruction behaves more deterministically, whereas higher compression introduces stochasticity, enabling more flexible reconstruction of plausible inputs. \textit{Best viewed when zoomed-in and in color}.
}
\label{fig:viz_diversity}
\end{center}
\vskip -0.1in
\end{figure*}

\section{Discussion}
\label{sec:discussion}

\myparagraph{Distribution-aware compression.}
Traditional image compression methods optimize the rate-distortion trade-off~\citep{shannon1959coding}, prioritizing compactness over input fidelity. 
Building on this, we also aim to capture the broader input distribution during compression, generating compact representations suitable for latent generative models.
This approach introduces an additional dimension to the trade-off, perception or distribution fidelity~\citep{blau2018perception}, which aligns more closely with the rate-distortion-perception framework~\citep{blau2019rethinking}.

\myparagraph{Iterative and stochastic decoding.}
A key question within the rate-distortion-perception trade-off is whether the iterative, stochastic nature of diffusion decoding offers advantages over traditional single-step, deterministic methods~\citep{kingma2013auto}. 
The strengths of diffusion~\citep{ho2020denoising} lie in its iterative process, which progressively refines the latent space for more accurate reconstructions, while stochasticity allows for capturing complex variations within the distribution. 
Although iterative methods may appear less efficient, our formulation is optimized to achieve optimal results in just three steps and also supports single-step decoding, ensuring decoding efficiency remains practical (see \cref{fig:ablation} (left)). 
While stochasticity might suggest the risk of ``hallucination'' in reconstructions, the outputs remain faithful to the underlying distribution by design, producing perceptually plausible results.
This advantage is particularly evident under extreme compression scenarios (see \cref{fig:viz_down}), with the degree of stochasticity adapting based on compression levels (see \cref{fig:viz_diversity}).

\myparagraph{Multi-step \textit{vs.}\ single-step decoding.} While replacing single-step decoding with an iterative process may seem counter-intuitive due to increased computational cost, the diffusion-based decoder addresses this concern in three key ways. First, it offers scalable inference, where even a single-step variant already outperforms a plain VAE decoder, and additional steps further enhance quality (see \cref{tbl:one-step}). Second, it provides controllable trade-offs between computation and visual fidelity, allowing the number of steps to be adjusted at inference time based on application needs. Third, as shown in \cref{tbl:cond-ldm}, it enables training under higher compression ratios, which helps offset the added cost of iterative decoding by reducing the size of latent representations.

\myparagraph{Scalability.}
As discussed in \cref{sec:exp:reconstruction}, our diffusion-based decoding method maintains the resolution generalizability typically found in standard autoencoders. This feature is highly practical: the autoencoder is trained on lower-resolution images, while the subsequent latent generative model is trained on latents derived from higher-resolution inputs. 
However, we acknowledge that memory overhead and throughput become concerns with our UNet-based diffusion decoder, especially for high-resolution inputs. 
This challenge becomes more pronounced as models, datasets, or resolutions scale up.
A promising future direction is patch-based diffusion~\citep{ding2023patched,wang2024patch}, which partitions the input into smaller, independently processed patches. This approach has the potential to reduce memory usage and enable faster parallel decoding. \section{Related work}
\label{sec:relatedwork}

\myparagraph{Image tokenization.}
Image tokenization is crucial for effective generative modeling, transforming images into compact, structured representations.
A common approach employs an autoencoder framework~\citep{hinton2006reducing}, where the encoder compresses images into low-dimensional latent representations, and the decoder reconstructs the original input.
These latent representations can be either discrete commonly used in autoregressive models~\citep{van2016conditional,van2017neural,chen2020generative,chang2022maskgit,yu2023magvit,kondratyuk2024videopoet}, or continuous, as found in diffusion models~\citep{ho2020denoising,dhariwal2021diffusion,rombach2022high,peebles2023scalable,gupta2023photorealistic,videoworldsimulators2024}.
The foundational form of visual autoencoding today originates from \citet{van2017neural}.
While advancements have been made in modeling~\citep{yu2022vector,yu2024image,chen2024deep}, objectives~\citep{zhang2018unreasonable,karras2019style,esser2021taming}, and quantization methods~\citep{yu2023language,zhao2024image}, the core encoding-and-decoding scheme remains largely the same.

In this work, we propose a new perspective by replacing the traditional decoder with a diffusion process.
Specifically, our new formulation retains the encoder but introduces a conditional diffusion decoder.
Within this framework, we systematically study various design choices, resulting in a significantly enhanced autoencoding setup.

Additionally, we refer to the recent work MAR~\citep{li2024autoregressive}, which leverages diffusion to model per-token distribution in autoregressive frameworks.
In contrast, our approach models the overall input distribution in autoencoders using diffusion.
This difference leads to distinct applications of diffusion during generation. 
For instance, MAR generates samples autoregressively, decoding each token iteratively using diffusion, token by token.
In our method, we first sample all tokens from the downstream generative model and then decode them iteratively using diffusion as a whole.

\myparagraph{Image compression.} 
Our work shares similarities with recent image compression approaches that leverage diffusion models. 
For example, \citet{hoogeboom2023high,birodkar2024sample} use diffusion to refine autoencoder residuals, enhancing high-frequency details. 
\citet{yang2024lossy} employs a diffusion decoder conditioned on quantized discrete codes and omits the GAN loss. 
However, these methods primarily focus on the traditional rate-distortion tradeoff, balancing rate (compactness) and distortion (input fidelity) \citep{shannon1959coding}, with the goal of storing and transmitting data efficiently without significant loss of information.

In this work, we emphasize perception (distribution fidelity) alongside the rate-distortion tradeoff, ensuring that reconstructions more closely align with the overall data distribution~\citep{heusel2017gans,zhang2018unreasonable,blau2019rethinking}, thereby enhancing the decoded results from the sampled latents of downstream generative models.
We achieve this by directly integrating the diffusion process into the decoder, unlike \citet{hoogeboom2023high,birodkar2024sample}. Moreover, unlike \citet{yang2024lossy}, we do not impose strict rate-distortion regularization in the latent space and allow the GAN loss to synergize with our approach.

\myparagraph{Diffusion decoder.}
Several studies~\citep{preechakul2022diffusion,shi2022divae,pernias2023wurstchen,nguyen2024swiftbrush,sauer2025adversarial,luo2023latent} have explored diffusion decoders conditioned on compressed latents of the input, which are relevant to our work. We outline the key differences between these works and \OURS: First, prior works have not fully leveraged the synergy between diffusion decoders and standard VAE training objectives. In this work, we enhance state-of-the-art VAE objectives by replacing the reconstruction loss with a score matching loss and adapting LPIPS and GAN losses to ensure compatibility with diffusion decoders. These changes yield significant improvements in autoencoding performance, as evidenced by lower rFID scores and faster inference. Second, we are the first to investigate various parameterizations (\eg, epsilon and velocity) and demonstrate that modern velocity parameterization, coupled with optimized train and test-time noise scheduling, provides substantial benefits. These enhancements improve both reconstruction performance and sampling efficiency. Third, previous diffusion-based decoders~\citep{preechakul2022diffusion,shi2022divae,pernias2023wurstchen}, which often rely on ad-hoc techniques like distillation or consistency regularization to speed up inference~\citep{nguyen2024swiftbrush,sauer2025adversarial,luo2023latent}, our approach achieves fast decoding (1 to 3 steps) without such techniques. This is made possible by integrating our proposed objectives and parameterizations. Last but not least, \OURS exhibits strong resolution generalization capabilities, a key property of standard VAEs. In contrast, models like DiffusionAE~\citep{preechakul2022diffusion} and DiVAE~\citep{shi2022divae} either lack this ability or are inherently limited. For example, DiVAE's bottleneck add/concat design restricts its capacity to generalize across resolutions.

SWYCC~\citep{birodkar2024sample} also explores joint learning of continuous encoders and decoders using a diffusion model. However, SWYCC differs fundamentally from our approach: it replaces the GAN loss with a diffusion-based loss, while we focus on identifying optimal synergies between traditional autoencoding losses (including GAN loss) and diffusion-based decoding. Our goal is to identify an optimal strategy for combining these elements, rather than simply substituting one for another.

Another closely related work, DiTo~\citep{chen2025diffusion}, also presents a diffusion-based tokenizer which learns compact visual representations for image generation. Its main insight is that a single diffusion learning objective is capable of training scalable image tokenizers. More than that, our method demonstrates that traditional autoencoding losses such as LPIPS and GAN losses are complimentary to the diffusion target, leading to better reconstruction quality. This design substantially differ our work from DiTo.

While following a different motivation, \citet{lee2023minimizing} essentially also proposes a VAE with a denoising decoder but uses the encoding as the ``initial noise'' instead of as conditioning for a standard diffusion model starting from a standard Gaussian distribution. This idea could be potentially used for speeding up the proposed approach, which we will explore in the future.

\myparagraph{Image generation.}
Recent advances in image generation span a wide range of approaches, including VAEs~\citep{kingma2013auto}, GANs~\citep{goodfellow2014generative}, autoregressive models~\citep{chen2020generative} and diffusion models~\citep{song2020score,ho2020denoising}.
Among these, diffusion models have emerged as the leading approach for generating high-dimensional data such as images~\citep{saharia2022palette,baldridge2024imagen,esser2024scaling} and videos~\citep{videoworldsimulators2024,gupta2023photorealistic}, where the gradual refinement of global structure is crucial. The current focus in diffusion-based generative models lies in advancing architectures~\citep{rombach2022high,peebles2023scalable,hoogeboom2023simple}, parameterizations~\citep{karras2022elucidating,kingma2024understanding,ma2024sit,esser2024scaling}, or better training dynamics~\citep{nichol2021improved,chen2023importance,chen2023generalist}.
However, tokenization, an essential component in modern diffusion models, often receives less attention.

In this work, we focus on providing compact continuous latents without applying quantization during autoencoder training~\citep{rombach2022high}, as they have been shown to be effective in state-of-the-art latent diffusion models~\citep{rombach2022high,saharia2022palette,peebles2023scalable,esser2024scaling,baldridge2024imagen}.
We compare our autoencoding performance against the baseline approach~\citep{esser2021taming} using the DiT framework~\citep{peebles2023scalable} as the downstream generative model. \section{Experiment setups}
\label{app:exp}

In this section, we provide additional details on our experiment configurations for reproducibility.

\subsection{Model specifications}
\label{app:exp:spec}

\cref{tbl:spec} summarizes the primary architecture details for each decoder variant. The channel dimension is the number of channels of the first U-Net layer, while the depth multipliers are the multipliers for subsequent resolutions. The number of residual blocks denotes the number of residual stacks contained in each resolution.

\begin{table}[t]
\caption{\textbf{Hyper-parameters for decoder variants.}}
\label{tbl:spec}
\vskip 0.1in
\begin{center}
\resizebox{0.98\linewidth}{!}{\begin{tabular}{l|ccc}
\toprule
\textbf{Model} & \textbf{Channel dim.} & \textbf{Depth multipliers} & \textbf{\# of blocks} \\
\midrule
Base (B) & 64 & \{1, 1, 2, 2, 4\} & 2 \\
Medium (M) & 96 & \{1, 1, 2, 2, 4\} & 2 \\
Large (L) & 128 & \{1, 1, 2, 2, 4\} & 2 \\
Extra-large (XL) & 128 & \{1, 1, 2, 2, 4\} & 4 \\
Huge (H) & 256 & \{1, 1, 2, 2, 4\} & 2 \\
\bottomrule
\end{tabular}
}
\end{center}
\vskip -0.1in
\end{table}

\subsection{Implementation details}
\label{app:exp:imp}

During the training of discriminators, \citet{esser2021taming} introduced an adaptive weighting strategy for $\lambda_\text{adv}$. However, we notice that this adaptive weighting does not introduce any benefit which is consistent with the observation made by \citet{sadat2024litevae}. Thus, we set $\lambda_\text{adv} = 0.5$ in the experiments for more stable model training across different configurations.

The autoencoder loss follows \Eqref{eq:vqgan}, with weights set to $\lambda_\text{LPIPS} = 0.5$ and $\lambda_\text{adv} = 0.5$.
We use the Adam optimizer~\citep{kingma2015adam} with $\beta_1 = 0$ and $\beta_2 = 0.999$, applying a linear learning rate warmup over the first 5,000 steps, followed by a constant rate of 0.0001 for a total of one million steps.
The batch size is 256, with data augmentations including random cropping and horizontal flipping.
An exponential moving average of model weights is maintained with a decay rate of 0.999.
All models are implemented in JAX/Flax~\citep{jax2018github,flax2020github} and trained on TPU-v5lite pods.

\subsection{Latent diffusion models}
\label{app:exp:ldm}

We follow the setting in~\citet{peebles2023scalable} to train the latent diffusion models for unconditional image generation on the ImageNet dataset. The DiT-XL/2 architecture is used for all experiments. The diffusion hyperparameters from ADM~\citep{dhariwal2021diffusion} are kept. To be specific, we use a $t_\text{max} = 1000$ linear variance schedule ranging from 0.0001 to 0.02, and results are generated using 250 DDPM sampling steps. For simplicity and training stability, we remove the variational lower bound loss term during training, which leads to a slight drop in generation qualities.

All models are trained with Adam~\citep{kingma2015adam} with no weight decay. We use a constant learning rate of 0.0001 and a batch size of 256. Horizontal flipping and random cropping are used for data augmentation. We maintain an exponential moving average of DiT weights over training with a decay of 0.9999. We use identical training hyperparameters across all experiments and train models for one million steps in total. No classifier-free guidance~\citep{ho2022classifier} is employed in all the experiments. Inference throughputs are computed on a Tesla H100 GPU. \section{Additional experimental results}
\label{app:exp:additional}

We note that all experiments conducted in this section are under the \OURS-lite configuration.

\begin{table}[t]
\caption{\textbf{Image reconstruction results on ImageNet $128 \times 128$.}}
\label{tbl:addition}
\vskip 0.1in
\begin{center}
\resizebox{0.98\linewidth}{!}{\begin{tabular}{l|cc}
\toprule
\textbf{Configuration} & \textbf{NFE}$\downarrow$ & \textbf{rFID}$\downarrow$ \\
\midrule
\textit{Baseline (c)} in \cref{tbl:ablation}: \\
~Inject conditioning by channel-wise concatenation & 50 & 22.04 \\
~Inject conditioning by AdaGN & 50 & 22.01 \\
\midrule
\textit{Baseline (e)} in \cref{tbl:ablation}: \\
~Matching the distribution of \( \hat{\vx}_0^t \) and \( \vx_0 \) & - & N/A \\
~Matching the trajectory of $\vx_{t} \rightarrow \vx_0$ & 5 & 8.24 \\
~Matching the trajectory of $\vx_{t} \rightarrow \vx_{t - \Delta t}$ & 5 & 10.53 \\
\bottomrule
\end{tabular}
}
\end{center}
\vskip -0.1in
\end{table}

\begin{table*}[t]
\caption{
\textbf{Model scaling and resolution generalization analysis.} 
Five model variants are trained and evaluated.
$\Delta_\text{rFID}$ represents the absolute differences (or relative ratio) in rFID  between the corresponding model size variants of VAE and \OURS. 
$^\dag$ denotes resolution generalization experiments.
To fairly evaluate the impact of \OURS under controlled model parameters, we highlight three groups of model variants with comparable parameters, using different colors.
}
\label{tbl:reconstruction}
\vskip 0.1in
\begin{center}
\resizebox{0.76\textwidth}{!}{\begin{tabular}{lc|cc|cc|cc}
\toprule
\multirow{2}[2]{*}{\textbf{Model}} & \multirow{2}[2]{*}{\textbf{$\gG$ params (M)}} & \multicolumn{2}{c|}{\textbf{ImageNet $128 \times 128$}} & \multicolumn{2}{c|}{\textbf{ImageNet $256 \times 256$ $^\dag$}} & \multicolumn{2}{c}{\textbf{ImageNet $512 \times 512$ $^\dag$}} \\
\cmidrule(lr){3-4} \cmidrule(lr){5-6} \cmidrule(lr){7-8}
& & \textbf{rFID}$\downarrow$ & $\Delta_\text{rFID}$ & \textbf{rFID}$\downarrow$ & $\Delta_\text{rFID}$ & \textbf{rFID}$\downarrow$ & $\Delta_\text{rFID}$ \\
\midrule
VAE (B) & 10.14 & 11.15 & - & 5.74 & - & 3.69 & - \\
\bluecell VAE (M) & \bluecell 22.79 & \bluecell 9.26 & \bluecell - & \bluecell 4.63 & \bluecell - & \bluecell 2.69 & \bluecell - \\
\redcell VAE (L) & \redcell 40.48 & \redcell 8.49 & \redcell - & \redcell 4.78 & \redcell - & \redcell 2.78 & \redcell - \\
VAE (XL) & 65.27 & 7.58 & - & 4.42 & - & 2.41 & - \\
\greencell VAE (H) & \greencell 161.81 & \greencell 7.12 & \greencell - & \greencell 4.29 & \greencell - & \greencell 2.37 & \greencell - \\
\midrule
\bluecell \OURS (B) & \bluecell 20.63 & \bluecell 6.24 & \bluecell 4.91 (44.0$\%$) & \bluecell 3.90 & \bluecell 1.84 (32.0$\%$) & \bluecell 2.06 & \bluecell 1.63 (44.2$\%$) \\
\redcell \OURS (M) & \redcell 49.33 & \redcell 5.42 & \redcell 3.84 (41.5$\%$) & \redcell 2.79 & \redcell 1.84 (39.7$\%$) & \redcell 2.02 & \redcell 0.67 (24.9$\%$) \\
\OURS (L) & 88.98 & 4.71 & 3.78 (44.5$\%$) & 2.60 & 2.03 (43.8$\%$) & 1.92 & 0.86 (30.9$\%$) \\
\greencell \OURS (XL) & \greencell 140.63 & \greencell 4.18 & \greencell 3.40 (44.9$\%$) & \greencell 2.38 & \greencell 2.04 (46.2$\%$) & \greencell 1.82 & \greencell 0.59 (24.5$\%$) \\
\OURS (H) & 355.62 & 4.04 & 3.08 (43.3$\%$) & 2.31 & 1.98 (46.2$\%$) & 1.78 & 0.59 (24.9$\%$) \\
\bottomrule
\end{tabular}
}
\end{center}
\vskip -0.1in
\end{table*}

\begin{table*}[t]
\caption{
\textbf{Unconditional image generation quality.} 
The DiT-XL/2 is trained on latents provided by the trained autoencoders, VAE and \OURS, with varying model sizes using ImageNet.
We evaluate the generation quality at resolutions of $128 \times 128$ and $256 \times 256$ using four standard metrics.
Additionally, we report rFID to determine if the improvement trend observed in reconstruction task extends to the generation task.
We highlight three groups of model variants with comparable parameters.
}
\label{tbl:generation}
\vskip 0.1in
\begin{center}
\resizebox{0.7\textwidth}{!}{\begin{tabular}{l|ccccc|ccccc}
\toprule
\multirow{2}[2]{*}{\textbf{Model}} & \multicolumn{5}{c|}{\textbf{ImageNet $128 \times 128$}} & \multicolumn{5}{c}{\textbf{ImageNet $256 \times 256$}} \\
\cmidrule(lr){2-6} \cmidrule(lr){7-11}& \textbf{rFID}$\downarrow$  &  \textbf{FID}$\downarrow$ & \textbf{IS}$\uparrow$ & \textbf{Prec.}$\uparrow$ & \textbf{Rec.}$\uparrow$ & \textbf{rFID}$\downarrow$ & \textbf{FID}$\downarrow$ & \textbf{IS}$\uparrow$ & \textbf{Prec.}$\uparrow$ & \textbf{Rec.}$\uparrow$ \\
\midrule
VAE (B) & \textcolor{lightgray}{11.15} & 36.8 & 17.9 & 0.48 & 0.53 & \textcolor{lightgray}{5.74} & 46.6 & 23.4 & 0.45 & 0.56 \\
\bluecell VAE (M) & \bluecell \textcolor{lightgray}{9.26} & \bluecell 34.6 & \bluecell 18.2 & \bluecell 0.49 & \bluecell 0.55 & \bluecell \textcolor{lightgray}{4.63} & \bluecell 44.7 & \bluecell 23.8 & \bluecell 0.47 & \bluecell 0.58 \\
\redcell VAE (L) & \redcell \textcolor{lightgray}{8.49} & \redcell 33.9 & \redcell 18.4 & \redcell 0.50 & \redcell 0.56 & \redcell \textcolor{lightgray}{4.78} & \redcell 44.3 & \redcell 24.7 & \redcell 0.47 & \redcell 0.59 \\
VAE (XL) & \textcolor{lightgray}{7.58} & 31.7 & 19.3 & 0.51 & 0.57 & \textcolor{lightgray}{4.42} & 43.1 & 24.9 & 0.47 & 0.59 \\
\greencell VAE (H) & \greencell \textcolor{lightgray}{7.12} & \greencell 30.9 & \greencell 19.8 & \greencell 0.52 & \greencell 0.57 & \greencell \textcolor{lightgray}{4.29} & \greencell 41.6 & \greencell 25.9 & \greencell 0.48 & \greencell 0.59 \\
\midrule
\bluecell \OURS (B) & \bluecell \textcolor{lightgray}{6.24} & \bluecell 29.5 & \bluecell 20.7 & \bluecell 0.53 & \bluecell 0.59 & \bluecell \textcolor{lightgray}{3.90} & \bluecell 39.5 & \bluecell 25.2 & \bluecell 0.46 & \bluecell 0.61 \\
\redcell \OURS (M) & \redcell \textcolor{lightgray}{5.42} & \redcell 27.6 & \redcell 21.2 & \redcell 0.55 & \redcell 0.59 & \redcell \textcolor{lightgray}{2.79} & \redcell 35.4 & \redcell 26.2 & \redcell 0.51 & \redcell 0.62 \\
\OURS (L) & \textcolor{lightgray}{4.71} & 27.3 & 22.1 & 0.55 & 0.59 & \textcolor{lightgray}{2.60} & 34.8 & 26.5 & 0.51 & 0.63 \\
\greencell \OURS (XL) & \greencell \textcolor{lightgray}{4.18} & \greencell 25.3 & \greencell 22.7 & \greencell 0.55 & \greencell 0.59 & \greencell \textcolor{lightgray}{2.38} & \greencell 34.0 & \greencell 27.4 & \greencell 0.53 & \greencell 0.63 \\
\OURS (H) & \textcolor{lightgray}{4.04} & 24.9 & 23.0 & 0.56 & 0.60 & \textcolor{lightgray}{2.31} & 33.2 & 27.5 & 0.54 & 0.64 \\
\bottomrule
\end{tabular}
}
\end{center}
\vskip -0.1in
\end{table*}

\myparagraph{Conditioning.} In addition to injecting conditioning via channel-wise concatenation, we explore providing conditioning to the diffusion model by adaptive group normalization (AdaGN)~\citep{nichol2021improved,dhariwal2021diffusion}. To achieve this, we resize the conditioning (\ie, encoded latents) via bilinear sampling to the desired resolution of each stage in the U-Net model, and incorporates it into each residual block after a group normalization operation~\citep{wu2018group}. This is similar to adaptive instance norm~\citep{karras2019style} and FiLM~\citep{perez2018film}. We report the results in \cref{tbl:addition} (top), where we find that channel-wise concatenation and AdaGN obtain similar reconstruction quality in terms of rFID. Because of the additional computational cost required by AdaGN, we thus apply channel-wise concatenation in our model by default.

\begin{figure*}[p]
\vskip 0.1in
\begin{center}
\includegraphics[width=\linewidth]{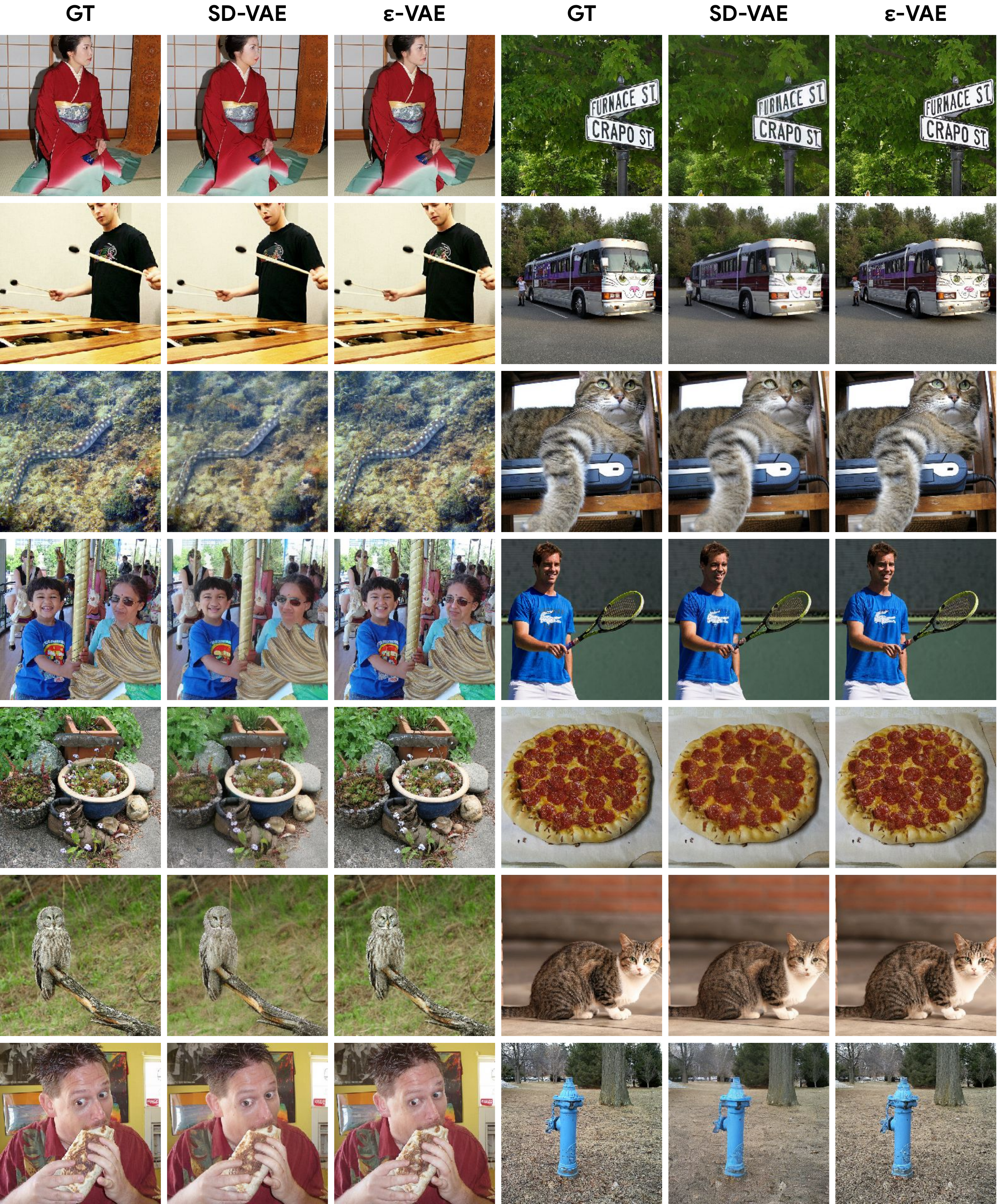}
\vskip -0.1in
\caption{
\textbf{Image reconstruction results under the SD-VAE configuration~\citep{rombach2022high} at the resolution of $256 \times 256$.} \OURS produces significantly better visual details than SD-VAE when reconstructing local regions with complex textures or structures, such as human faces and small texts. 
\textit{Best viewed when zoomed-in and in color}.
}
\label{fig:viz_sd-vae-256}
\end{center}
\vskip -0.1in
\end{figure*}

\myparagraph{Trajectory matching.} The proposed denoising trajectory matching objective matches the start-to-end trajectory $\vx_{t} \rightarrow \vx_0$ by default. One alternative choice is to directly matching the distribution of \( \hat{\vx}_0^t \) and \( \vx_0 \) without coupling on $\vx_t$. However, we find this formulation leads to unstable training and could not produce reasonable results. Here, we present the results when matching the trajectory of $\vx_{t} \rightarrow \vx_{t - \Delta t}$, which is commonly used in previous work~\citep{xiao2021tackling,wang2024phased}. Specifically, for each timestep $t$ during training, we randomly sample a step $\Delta t$ from $(0, t)$. Then, we construct the real trajectory by computing $\vx_{t - \Delta t}$ via \Eqref{eq:rf_traj} and concatenating it with $\vx_{t}$, while the fake trajectory is obtained in a similar way but using \Eqref{eq:rf_est} instead. \cref{tbl:addition} (bottom) shows the comparison. We observe that matching trajectory $\vx_{t} \rightarrow \vx_0$ yields better performance than matching trajectory $\vx_{t} \rightarrow \vx_{t - \Delta t}$, confirming the effectiveness of the proposed objective which is designed for the rectified flow formulation.

\myparagraph{Comparisons with plain diffusion ADM.} Under the same training setup of \cref{tbl:ablation}, we directly trained a plain diffusion model (ADM) for comparison, which resulted in rFID score of 38.26. Its conditional form is already provided as a baseline in \cref{tbl:ablation}, achieving 28.22. This demonstrates that our conditional form $p(\vx_{t-1}|\vx_t,\vz)$ offers a better approximation of the true posterior $q(\vx_{t-1}|\vx_t,\vx_0)$ compared to the standard form $p(\vx_{t-1}|\vx_t)$. By further combining LPIPS and GAN loss, we achieve rFID of 8.24, outperforming its VAE counterpart, which achieves 11.15. With better training configurations, our final rFID improves to 6.24. This progression, from plain diffusion ADM to \OURS, underscores the significance of our proposals and their impact.

\myparagraph{Model scaling.} We investigate the impact of model scaling by comparing VAE and \OURS across five model variants, all trained and evaluated at a resolution of $128 \times 128$, as summarized in \cref{tbl:reconstruction}.
The results demonstrate that \OURS consistently achieves significantly better rFID scores than VAE, with an average relative improvement of over 40$\%$, and even the smallest \OURS model outperforms VAE at largest scale.
While the U-Net-based decoder of \OURS has about twice as many parameters as standard decoder of VAE, grouping models by similar sizes, highlighted in different colors, shows that performance gains are not simply due to increased model parameters.

\cref{tbl:generation} presents the unconditional image generation results of VAE and \OURS at resolutions of $128 \times 128$ and $256 \times 256$. In addition to FID, we report Inception Score (IS) \citep{salimans2016improved} and Precision/Recall~\citep{kynkaanniemi2019improved} as secondary metrics.
The results show that \OURS consistently outperforms VAE across all model scales.
Notably, \OURS (B) surpasses VAE (H), consistent with our earlier findings in \Secref{sec:exp:reconstruction}.
These results further demonstrate the effectiveness of \OURS from the generation perspective.

\myparagraph{Results with classifier-free
guidance.} We provide additional results with classifier-free
guidance~\citep{ho2022classifier} under the $8 \times 8$ downsample factor in \cref{tbl:cond-ldm-cfg}. We find that \OURS (M) performs relatively 20\% better than SD-VAE and further improvements are obtained after we scale up our model to \OURS (H). These results are consistent with the results without classifier-free guidance in \cref{tbl:cond-ldm}, confirming the effectiveness of our model.

\begin{table}[t]
\caption{\textbf{Benchmarking class-conditional image generation on ImageNet $256 \times 256$.} We use the DiT-XL/2 architecture~\citep{esser2024scaling} for latent diffusion models and apply classifier-free guidance~\citep{ho2022classifier}.
}
\label{tbl:cond-ldm-cfg}
\vskip 0.1in
\begin{center}
\resizebox{0.84\linewidth}{!}{\begin{tabular}{c|l|cc}
\toprule
\multirowcell{2}{\textbf{Downsample}\\ \textbf{factor}} & \multirow{2}{*}{\textbf{Method}} & \multirow{2}{*}{\textbf{FID}$\downarrow$} \\
& \\
\midrule
\multirow{3}{*}{$8 \times 8$} & SD-VAE~\citep{rombach2022high} & 3.51 \\
& \bluecell \OURS-SD (M) & \bluecell \underline{2.83} \\
& \bluecell \OURS-SD (H) & \bluecell \textbf{2.69} \\
\bottomrule
\end{tabular}
}
\end{center}
\vskip -0.1in
\end{table}

\section{Additional visual results}
\label{app:visual}

We provide additional visual comparisons between \OURS and SD-VAE at the resolution of $256 \times 256$ (\cref{fig:viz_sd-vae-256}). Our observations indicate that \OURS delivers significantly better visual quality than SD-VAE, particularly when reconstructing local regions with complex textures or structures, such as human faces and small text.

\end{document}